\documentclass{article}

\usepackage{microtype}
\usepackage{graphicx}
\usepackage{subfigure}
\usepackage{booktabs} 
\usepackage{hyperref}
\usepackage{url}
\usepackage{booktabs}       
\usepackage{amsfonts}     
\usepackage{nicefrac}       
\usepackage{microtype}      
\usepackage{xcolor}         
\usepackage{graphicx}
\usepackage{wrapfig}
\usepackage{multirow}
\usepackage{epsfig}
\usepackage{amsmath}
\usepackage{amssymb}
\usepackage{subfigure}
\usepackage{marvosym}
\usepackage{color}
\usepackage{threeparttable}
\usepackage{algorithm}
\usepackage{setspace}
\usepackage{amsthm}
\usepackage{verbatim}
\usepackage{calligra}
\usepackage{algorithm}
\usepackage{algorithmicx}
\usepackage{algpseudocode} 
\usepackage{amsmath} 
\usepackage{balance}
\newcommand{\update}[1]{{\textcolor{black}{#1}}}

\renewcommand{\algorithmicrequire}{\textbf{Input:}}
\renewcommand{\algorithmicensure}{\textbf{Return:}}

\usepackage{hyperref}

\usepackage[accepted]{icml2024}

\usepackage{amsmath}
\usepackage{amssymb}
\usepackage{mathtools}
\usepackage{amsthm}

\usepackage[capitalize,noabbrev]{cleveref}

\theoremstyle{plain}

\theoremstyle{definition}

\theoremstyle{remark}

\usepackage[textsize=tiny]{todonotes}

\icmltitlerunning{Submission and Formatting Instructions for ICML 2024}

\begin{document}

\twocolumn[
\icmltitle{PDETime: Rethinking Long-Term Multivariate Time Series Forecasting from the perspective of partial differential equations}



\icmlsetsymbol{equal}{*}

\begin{icmlauthorlist}
\icmlauthor{Shiyi Qi}{loc:a}
\icmlauthor{Zenglin Xu}{loc:a}
\icmlauthor{Yiduo Li}{loc:a}
\icmlauthor{Liangjian Wen}{loc:b}
\icmlauthor{Qingsong Wen}{loc:c}
\icmlauthor{Qifan Wang}{loc:d}
\icmlauthor{Yuan Qi}{loc:e}
\end{icmlauthorlist}

\icmlaffiliation{loc:a}{Harbin Institute of Technology, Shenzhen}
\icmlaffiliation{loc:b}{Huawei Technologies Ltd}
\icmlaffiliation{loc:c}{Squirrel AI}
\icmlaffiliation{loc:d}{Meta AI}
\icmlaffiliation{loc:e}{Fudan University}

\icmlcorrespondingauthor{Shiyi Qi}{21s051040@stu.hit.edu.cn}
\icmlcorrespondingauthor{Zenglin Xu}{zenglin@gmail.com}
\icmlcorrespondingauthor{Yiduo Li}{lisa\_liyiduo@163.com}
\icmlcorrespondingauthor{Liangjian Wen}{wenliangjian1@huawei.com}
\icmlcorrespondingauthor{Qingsong Wen}{qingsongedu@gmail.com}
\icmlcorrespondingauthor{Qifan Wang}{wqfcr@fb.com}
\icmlcorrespondingauthor{Yuan Qi}{qiyuan@fudan.edu.cn}
\icmlkeywords{Machine Learning, ICML}

\vskip 0.3in
]



\printAffiliationsAndNotice{\icmlEqualContribution} 

\begin{abstract}

Recent advancements in deep learning have led to the development of various models for long-term multivariate time-series forecasting (LMTF), many of which have shown promising results. Generally, the focus has been on historical-value-based models, which rely on past observations to predict future series. Notably, a new trend has emerged with time-index-based models, offering a more nuanced understanding of the continuous dynamics underlying time series.
Unlike these two types of models that aggregate the information of spatial domains or temporal domains, in this paper, we consider multivariate time series as spatiotemporal data regularly sampled from a continuous dynamical system, which can be represented by partial differential equations (PDEs), with the spatial domain being fixed. Building on this perspective, we present PDETime, 
a novel LMTF model inspired by the principles of Neural PDE solvers, following the encoding-integration-decoding operations. Our extensive experimentation across seven diverse real-world LMTF datasets reveals that PDETime not only adapts effectively to the intrinsic spatiotemporal nature of the data but also sets new benchmarks, achieving state-of-the-art results. 
\end{abstract}
\section{Introduction}
Multivariate time series forecasting plays a pivotal role in diverse applications such as weather prediction~\cite{angryk2020multivariate}, energy consumption~\cite{demirel2012forecasting}, healthcare~\cite{matsubara2014funnel}, and traffic flow estimation~\cite{li2017diffusion}. 
In the field of multivariate time series forecasting, two typical approaches have emerged: historical-value-based~\cite{zhou2021informer,wu2021autoformer, zeng2023transformers, nie2022time}, and time-index-based models~\cite{woo2023learning,naour2023time}. 
Historical-value-based models predict future time steps utilizing historical observations, $\hat{x}_{t+1}=f_{\theta}(x_{t}, x_{t-1},...)$, while time-index-based models only  use the corresponding time-index features, $\hat{x}_{t+1}=f_{\theta}(t+1)$. 
Historical-value-based models have been widely explored due to their simplicity and effectiveness, making them the state-of-the-art models in the field. 
However, it is important to acknowledge that multivariate time series data are often discretely sampled from continuous dynamical systems. This characteristic poses a challenge for historical-value-based models, as they tend to capture spurious correlations limited to the sampling frequency~\cite{gong2017causal, woo2023learning}.
Moreover, despite their widespread use and effectiveness, the progress in developing historical-value-based models has encountered bottlenecks~\cite{li2023revisiting}, making it challenging to further improve their performance. This limitation motivates the exploration of alternative approaches that can overcome the shortcomings of historical-value-based models and potentially deliver more robust forecasting models.

In recent years, deep time-index-based models have gradually been studied~\cite{woo2023learning,naour2023time}. These models intrinsically avoid the aforementioned problem of historical-value-based models, by mapping the time-index features to target predictions in continuous space through implicit neural representations~\cite{tancik2020fourier,sitzmann2020implicit} (INRs). 
However, time-index-based models are only characterized by time-index coordinates, which makes them ineffective in modeling complex time series data and lacking in exploratory capabilities.

In this work, we propose a novel perspective by treating multivariate time series as spatiotemporal data discretely sampled from a continuous dynamical system, described by partial differential equations (PDEs) with fixed spatial domains as described in Eq~\ref{eq:pde_def}. 

From the PDE perspective, we observe that historical-value-based and time-index-based models can be considered as simplified versions of Eq~\ref{eq:pde}, where either the temporal or spatial information is removed.

By examining Eq~\ref{eq:pde}, we can interpret historical-value-based models as implicit inverse problems. These models infer spatial domain information, such as the position and physical properties of sensors, from historical observations. By utilizing the initial condition and spatial domain information, these models predict future sequences. Mathematically, historical-value-based models can be formulated as:
\begin{align}
    \hat{x}_{\tau}=x_{\tau_{0}}+ \int_{\tau_{0}}^{\tau} f_{\theta}(x) d\tau
\end{align}
where $x$ represents the spatial domain information.

Time-index-based models, on the other hand, can be seen as models that consider only the temporal domain. These models solely rely on time-index coordinates for prediction and do not explicitly incorporate spatial information. In this case, the prediction is formulated as:
\begin{align}
    \hat{x}_{\tau}= \int_{\tau_{0}}^{\tau} f_{\theta}(t_{\tau}) d\tau
\end{align}

Motivated by the limitations of existing approaches and inspired by neural PDE solvers, we introduce PDETime, a PDE-based model for LMTF. PDETime considers LMTF as an Initial Value Problem, explicitly incorporating both spatial and temporal domains and requiring the specification of an initial condition. 
Specifically, PDETime first predicts the partial derivative term $E_{\theta}(x_{his},t_{\tau})=\alpha_{\tau} \approx\frac{\partial u(x,t_{\tau})}{\partial t_{\tau}}$ utilizing an encoder in latent space. Unlike traditional PDE problems, the spatial domain information of LMT is unknown. Therefore, the encoder utilizes historical observations to estimate the spatial domain information. Subsequently, a numerical solver is employed to calculate the integral term $z_{\tau}=\int_{\tau_{0}}^{\tau} \alpha_{\tau} d\tau$. The proposed solver effectively mitigates the error issue and enhances the stability of the prediction results compared to traditional ODE solvers~\cite{chen2018neural}. Finally, PDETime employs a decoder to map the integral term from the latent space to the spatial domains and predict the results as $\hat{x}_{\tau}=x_{\tau_{0}}+D_{\phi}(z_{\tau})$. Similar to time-index-based models, PDETime also utilizes meta-optimization to enhance its ability to extrapolate across the forecast horizon. Additionally, PDETime can be simplified into a historical-value-based or time-index-based model by discarding either the time or spatial domains, respectively.


In short, we summarize the key contributions of this work:

\begin{itemize}
    \item We propose a novel perspective for LMTF by considering it as spatiotemporal data regularly sampled from PDEs along the temporal domains.
    \item We introduce PDETime, a PDE-based model inspired by neural PDE solvers, which tackles LMTF as an Initial Value Problem of PDEs. PDETime incorporates encoding-integration-decoding operations and leverages meta-optimization to extrapolate future series.
    \item We extensively evaluate the proposed model on seven real-world benchmarks across multiple domains under the long-term setting. Our empirical studies demonstrate that PDETime consistently achieves state-of-the-art performance.
\end{itemize}
\section{Related Work}
\textbf{Multivariate Time Series Forecasting.} 
With the progressive breakthrough made in deep learning, deep models have been proposed to tackle various time series forecasting applications. Depending on whether temporal or spatial is utilized, these models are classified into historical-value-based~\cite{zhou2021informer, zhou2022fedformer, zeng2023transformers, nie2022time, zhang2022crossformer,liu2023itransformer, liu2022non, liu2022scinet, wu2022timesnet}, and time-index-based models~\cite{woo2023learning}. Historical-value-based models, predicting target time steps utilizing historical observations, have been extensively developed and made significant progress in which a large body of work that tries to apply Transformer to forecast long-term series in recent years~\cite{wen2023transformers}. 
Early works like Informer~\cite{zhou2021informer} and LongTrans~\cite{li2019enhancing} were focused on designing novel mechanism to reduce the complexity of the original attention mechanism, thus capturing long-term dependency to achieve better performance. Afterwards, efforts were made to extract better temporal features to enhance the performance of the model~\cite{wu2021autoformer,zhou2022fedformer}. 
Recent work~\cite{zeng2023transformers} has found that a single linear channel-independent model can outperform complex transformer-based models. Therefore, the very recent channel-independent models like PatchTST~\cite{nie2022time} and DLinear~\cite{zeng2023transformers} have become state-of-the-art. 
In contrast, time-index-based models~\cite{woo2023learning, fons2022hypertime,jiang2023nert,naour2023time} are a kind of coordinated-based models, mapping coordinates to values, which was represented by INRs. These models have received less attention and their performance still lags behind historical-value-based models.
PDETime, unlike previous works, considers multivariate time series as spatiotemporal data and approaches the prediction target sequences from the perspective of partial differential equations (PDEs).


\textbf{Implicit Neural Representations.} Implicit Neural Representations are the class of works representing signals as a continuous function parameterized by multi-layer perceptions (MLPs)~\cite{tancik2020fourier, sitzmann2020implicit} (instead of using the traditional discrete representation). These neural networks have been used to learn differentiable representations of various objects such as images~\cite{henzler2020learning}, shapes~\cite{liu2020dist,liu2019learning}, and textures~\cite{oechsle2019texture}. However, there is limited research on INRs for times series~\cite{fons2022hypertime,jiang2023nert,woo2023learning, naour2023time,jeong2022time}. And previous works mainly focused on time series generation and anomaly detection~\cite{fons2022hypertime, jeong2022time}. 
DeepTime~\cite{woo2023learning} is the work designed to learn a set of basis INR functions for forecasting, however, its performance is worse than historical-value-based models. In this work, we use INRs to represent spatial domains and temporal domains.


\textbf{Neural PDE Solvers.} Neural PDE solvers which are used for temporal PDEs, are laying the foundations of what is becoming both a rapidly growing and impactful area of research. These neural PDE solvers fall into two broad categories, \textit{neural operator methods} and \textit{autoregressive methods}. The neural operator methods~\cite{kovachki2021neural,li2020fourier,lu2021learning} treat the mapping from initial conditions to solutions as time $t$ as an input-output mapping learnable via supervised learning. For a given PDE and given initial conditions $u_{0}$, the neural operator $\mathcal{M}$ is trained to satisfy $\mathcal{M}(t, u_{0})=u(t)$. However, these methods are not designed to generalize to dynamics for out-of-distribution $t$. In contrast, the autoregressive methods~\cite{bar2019learning, greenfeld2019learning, hsieh2019learning, yin2022continuous, brandstetter2021message} solve the PDEs iteratively. The solution of autoregressive methods at time $t+ \Delta t$ as $u(t+ \Delta)=\mathcal{A}(u(t), \Delta t)$. In this work, We consider multivariate time series as data sampled from a continuous dynamical system according to a regular time discretization, which can be described by partial differential equations. For the given initial condition $x_{\tau_{0}}$, we use the numerous solvers (Euler solver) to simulate target time step $x_{\tau}$.


\section{Method}
\label{sec:approach}

\begin{figure}[!]
\vskip 0.2in
\begin{center}
\centerline{\includegraphics[width=\columnwidth]{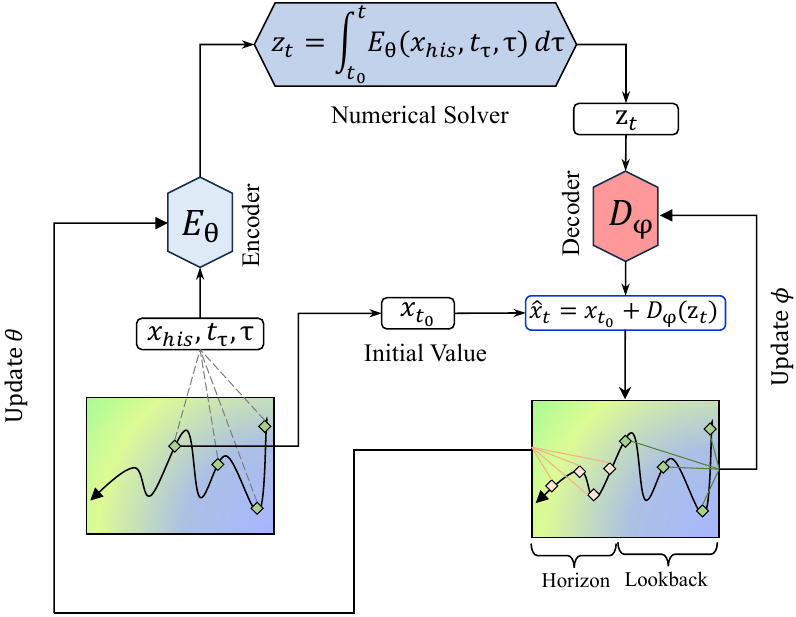}}\vspace{-3mm}
\caption{The framework of proposed PDETime which consists of an Encoder $E_{\theta}$, a Solver, and a Decoder $D_{\phi}$. Given the initial condition $x_{\tau_{0}}$, PDETime first simulates $\frac{\partial u(x,t_{\tau})}{\partial t_{\tau}}$ at each time step $t_{\tau}$ using the Encoder $E_{\theta}(x_{his},t_{\tau}, \tau)$; then uses the Solver to compute $\int_{\tau_{0}}^{\tau} \frac{\partial u(x,t_{\tau})}{\partial t_{\tau}} dt_{\tau} $, which is a numerical solver; finally, the Decoder maps integral term $z_{\tau}$ from latent space to special domains and predict the final results  $\hat{x}_{\tau}=x_{\tau_{0}}+D_{\phi}(z_{\tau})$. }
\label{fig:pdetime}
\end{center}\vspace{-3mm}
\vskip -0.2in
\end{figure}


\subsection{Problem Formulation}
In contrast to previous works~\cite{zhou2021informer, woo2023learning}, we regard multivariate time series as the spatiotemporal data regularly sampled from partial differential equations  along the temporal domain, which has the following form:
\begin{align}
    \mathcal{ F}(u, \frac{\partial u}{\partial t}, \frac{\partial u}{\partial x^{1}},..., \frac{\partial^{2} u}{\partial t^{2}},\frac{\partial^{2} u}{\partial x^{2}},...)=0, \nonumber \\
    u(x,t): \Omega \times \mathcal{T} \rightarrow \mathcal{V},
    \label{eq:pde_def}
\end{align}
subject to initial and boundary conditions. Here $u$ is a spatio-temporal dependent and multi-dimensional continuous vector field, $\Omega \in \mathbb{R}^{C}$ and $\mathcal{T} \in \mathbb{R}$ denote the spatial and temporal domains, respectively. For multivariate time series data, we regard channel attributes as spatial domains (e.g.,  the position and physical properties of sensors). 
The value of the spatial domains, $x$, is fixed but unknown (e.g., the position of sensors is fixed but unknown), while the value of the temporal domains is known.
We treat LMTF as the Initial Value Problems, where $u(x_{t},t) \in \mathbb{R}^{C}$ at any time $t$ is required to infer $u(x_{t'},t') \in \mathbb{R}^{C}$.


Consequently, we forecast the target series by the following formula: 
\begin{equation}
u(x,t)=u(x,t_{0})+\int_{t_{0}}^{t} \frac{\partial u(x,\tau)}{ \partial \tau} d\tau.
    \label{eq:pde}
\end{equation}

DETime initiates its process with a single initial condition, denoted as $u(x,t_{0})$, and leverages neural networks to project the system's dynamics forward in time. The procedure unfolds in three distinct steps. Firstly, PDETime generates a latent vector, $\alpha_{\tau}$ of a predefined dimension $d$, utilizing an encoder function, $E_{\theta}: \Omega \times \mathcal{T} \rightarrow \mathbb{R}^{d}$ (denoted as the ENC step). Subsequently, it employs an Euler solver, a numerical method, to approximate the integral term, $z_{t}=\int_{t_{0}}^{t} \alpha_{\tau} d\tau $, effectively capturing the system's evolution over time (denoted as the SOL step). In the final step, PDETime translates the latent vectors, 
$z_{t}$, back into the spatial domain using a decoder, 
 $D_{\phi}: \mathbb{R}^{d} \rightarrow \mathcal{V}$  to reconstruct the spatial function (denoted as the DEC step). This results in the following model, illustrated in Figure~\ref{fig:pdetime},
\begin{align}
    (ENC)\,\, \alpha_{\tau}&=E_{\theta}(x_{his}, t_{\tau},\tau)\\
    (SOL)\,\,\,\, z_{\tau}&=\int_{\tau_{0}}^{\tau} \alpha_{\tau} d\tau \\
    (DEC)\,\, x_{\tau}&=D_{\phi}(z_{\tau}) .
\end{align}
We describe the details of the components in Section~\ref{sec:component}.

\subsection{Components of PDETime}
\label{sec:component}

\subsubsection{Encoder: $\alpha_{\tau}=E_{\theta}(x_{his},t_{\tau},\tau)$}
\label{sec:forecaster}
\begin{figure}[!]
\vskip 0.2in
\begin{center}
\centerline{\includegraphics[width=0.9\columnwidth]{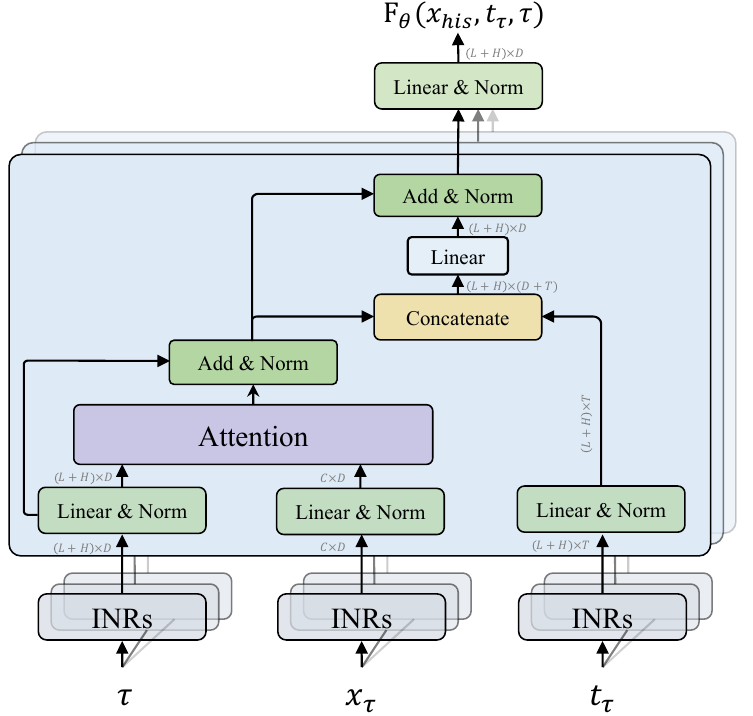}}\vspace{-4mm}
\caption{The architecture of the Encoder which is used to simulate $\frac{\partial u(x,t_{\tau})}{\partial t_{\tau}}$. In this work, we use INRs to represent $x_{his}$, $t_{\tau}$, and $\tau$, and then use the Attention mechanism and linear mapping to aggregate the information of spatial domains and temporal domains.}
\label{fig:forecaster}
\end{center}\vspace{-2mm}
\vskip -0.2in
\end{figure}

The Encoder computes the latent vector $\alpha_{\tau}$ representing  $\frac{\partial u(x,t_{\tau})}{\partial t_{\tau}}$. 
In contrast to previous works~\cite{chen2022crom, yin2022continuous} that compute 
$\frac{\partial u(x,t_{\tau})}{\partial t_{\tau}}$ using auto-differentiation powered by deep learning frameworks, which often suffers from poor stability due to uncontrolled temporal discretizations,
we propose directly simulating $\frac{\partial u(x,t_{\tau})}{\partial t_{\tau}}$ by a carefully designed neural networks $E_{\theta} (x_{his},t_{\tau},\tau)$. 
The overall architecture of $E_{\theta} (x_{his},t_{\tau},\tau)$ is shown in Figure~\ref{fig:forecaster}. As mentioned in Section~\ref{sec:approach}, since the spatial domains of MTS data are fixed and unknown, we can only utilize the historical data $x_{his} \in \mathbb{R}^{H \times C}$ to assist the neural network in fitting the value of spatial domains $x$. 
To represent the high frequency of $\tau$, $x_{his}$, and $t_{\tau}$, we use concatenated Fourier Features (CFF)~\cite{woo2023learning, tancik2020fourier} and SIREN~\cite{sitzmann2020implicit} with $k$ layers.
\begin{align}
    \tau^{(i)}&= \text{GELU}(W_{\tau}^{(i-1)}\tau^{(i-1)}+b_{\tau}^{(i-1)} )\nonumber \\
    x_{\tau}^{(i)}&= \sin(W_{x}^{(i-1)}x_{\tau}^{(i-1)}+b_{x}^{(i-1)}) \nonumber\\
    t_{\tau}^{(i)}&= \sin(W_{t}^{(i-1)}t_{\tau}^{(i-1)}+b_{t}^{(i-1)}), \,\,\, i=1,...,k 
\end{align}
where $x_{\tau}^{(0)} \in \mathbb{R}^{H \times C}=x_{his}$, $t_{\tau}^{(0)} \in \mathbb{R}^{t}$ is the temporal feature, and $\tau \in \mathbb{R}$ is the index feature where $\tau_{i}=\frac{i}{H+L}$ for $i=0,1,...,H+L$, $L$ and $H$ are the look-back and horizon length, respectively. We employ CFF to represent $\tau^{(0)}$, i.e. $\tau^{(0)}=[\sin(2\pi \textbf{B}_{1}\tau), \cos(2\pi \textbf{B}_{1}\tau),..., \sin(2 \pi \textbf{B}_{s} \tau), \cos (2 \pi \textbf{B}_{s} \tau)]$ where $\textbf{B}_{s} \in \mathbb{R}^{\frac{d}{2}\times c} $ is sampled from $\mathcal{N}(0,2^{s})$.
\par
After representing $\tau \in \mathbb{R}^{d}$, $t_{\tau} \in \mathbb{R}^{t}$, and $x_{\tau} \in \mathbb{R}^{d \times C}$ with INRs, the Encoder aggregates $x_{\tau}$ and $t_{\tau}$ using $\tau$, formulated as $f_{\theta}(\tau, t_{\tau},x_{\tau})$:
\begin{align}
    \tau&=\tau+\text{Attn}(\tau, x_{\tau}) \nonumber \\
    \tau&=\tau+\text{Linear}([\tau; t_{\tau}])
\end{align}
where $[\cdot;\cdot]$ is is the row-wise stacking operation. In this work, we utilize attention mechanism~\cite{vaswani2017attention} and linear mapping to aggregate the information of spatial domains and temporal domains, respectively. However, various networks can be designed to aggregate these two types of information.
\par
Unlike previous works~\cite{chen2022crom,yin2022continuous}, we do not directly make $\alpha_{\tau}=\frac{\partial u(x,t_{\tau})}{\partial t_{\tau}}$, but rather make $\int_{\tau_{0}}^{\tau}\alpha_{\tau} d \tau=\int_{\tau_{0}}^{\tau} \frac{\partial u(x, t_{\tau})}{\partial t_{\tau}} d \tau $ in latent space, without the need for autoregressive calculation of $\alpha_{t}$ which can effectively alleviate the error accumulation problem and make the prediction results more stable~\cite{brandstetter2021message}. 

The pseudocode of the Encoder of PDETime is summarized in Appendix~\ref{sec:code_Encoder}.

\subsubsection{Solver: $z_{t}=\int_{\tau_{0}}^{\tau} \alpha_{\tau}$}
\begin{figure}[!]
\vskip 0.2in
\begin{center}
\centerline{\includegraphics[width=0.9\columnwidth]{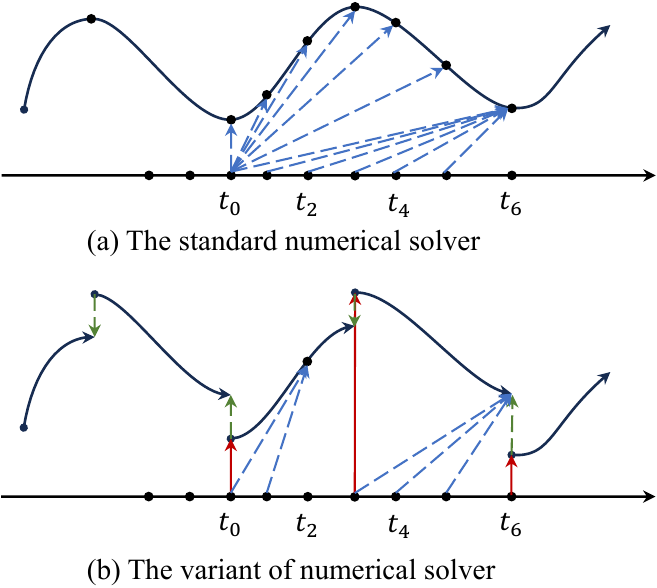}}\vspace{-3mm}
\caption{The framework of standard and variant of numerical solver proposed in our work.}
\label{fig:est}
\end{center}\vspace{-2mm}
\vskip -0.2in
\end{figure}

The Solver introduces a numerical solver (Euler Solver) to compute the integral term $z_{\tau}=\int_{\tau_{0}}^{\tau} \alpha_{\tau} d \tau$, which can be approximated as:
\begin{align}
    z_{\tau}&=\int_{\tau_{0}}^{\tau} \frac{\partial u(x_{\tau},t_{\tau})}{\partial t_{\tau}} d \tau \nonumber
    \approx \sum_{\tau=\tau_{0}}^{\tau}\frac{\partial u(x_{\tau}, t_{\tau})}{\partial t_{\tau}} \ast \Delta \tau \nonumber\\
    & \approx \sum_{\tau=\tau_{0}}^{\tau} E_{\theta}(x_{his},t_{\tau}, \tau) \ast \Delta \tau
    \label{eq:solver}
\end{align}
where we set $\Delta \tau=1$ for convenience. However, as shown in Figure~\ref{fig:est} (a), directly compute $z_{\tau}=\sum_{\tau=\tau_{0}}^{\tau} E_{\theta}(x_{his}, t_{\tau}, \tau) \ast \Delta \tau=\sum_{\tau=\tau_{0}}^{\tau} \alpha_{\tau} \ast \Delta \tau$ through Eq~\ref{eq:solver} can easily lead to error accumulation and gradient problems~\cite{rubanova2019latent,wu2022learning,brandstetter2021message}. To address these issues, we propose a modified approach, as illustrated in Figure~\ref{fig:est} (b). We divide the sequence into non-overlapping patches of length $S$, where $\frac{H+L}{S}$ patches are obtained. For $\tau(H+L) \mod S = 0$, we directly estimate the integral term as $z_{\tau} = f_{\psi}(x_{his},t_{\tau}, \tau)$ using a neural network $f_{\psi}$. Otherwise, we use the numerical solver to estimate the integral term with the integration lower limit $\lfloor \frac{\tau \cdot (H+L)}{S} \rfloor \cdot S$. This modification results in the following formula for the numerical solver:
\begin{align}
    z_{t}&=f_{\psi}(x_{\tau'},t_{\tau'}, \tau')+\int_{\tau'}^{\tau}f_{\theta}(x_{\tau},t_{\tau}, \tau) d \tau,  \nonumber\\
    \tau'& =\lfloor \frac{\tau * (H+L)}{s} \rfloor \ast S
    \label{eq:new_numerical_solver}
\end{align}
Furthermore, as shown in Figure~\ref{fig:est} (b), Eq~\ref{eq:new_numerical_solver} breaks the continuity. To address this, we introduce an additional objective function to ensure continuity as much as possible:
\begin{align}
    \mathcal{L}_{c}=\mathcal{L}(f_{\psi}(x_{\tau}, t_{\tau}, \tau),f_{\psi}(x_{\tau'}, t_{\tau'}, \tau')+\int_{\tau'}^{\tau}f_{\theta}(x_{\tau}, t_{\tau}, \tau) d \tau) \nonumber\\
    \mathrm{s.t.}\; \tau \ast(H+L) \mod S=0, \; \tau'=\frac{\tau\ast(L+H)-S}{(H+L)}.
\end{align}
The experimental results have shown that incorporating $\mathcal{L}_{c}$ improves the stability of PDETime. 

The pseudocode of the Solver of PDETime is summarized in Appendix~\ref{sec:code_Solver}.

\subsubsection{Decoder: $x_{\tau}=D_{\phi}(z_{\tau})+x_{\tau_{0}}$}
After estimating the integral term in latent space, we utilize a Decoder $D_{\phi}$ to decode $z_{\tau}$ into the spatial function. To tackle the distribution shift problem, we introduce a meta-optimization framework~\cite{woo2023learning,bertinetto2018meta} to enhance the extrapolation capability of PDETime. Specifically, we split the time series into look-back window $\textbf{X}_{t-L:t}=[x_{t-L},...,x_{t}] \in \mathbb{R}^{L \times C}$ and horizon window $\textbf{Y}_{t+1:t+H}=[x_{t+1},...,x_{t+H}] \in \mathbb{R}^{H \times C}$ pairs. We then use the parameters $\phi$ and $\theta$ to adapt the look-back window and horizon window through a bi-level problem:
\begin{align}
    \theta^{\ast}=\arg \min_{\theta} \sum_{\tau=\tau_{t+1}}^{\tau_{t+H}} \mathcal{L}_{p} ( D_{\phi}( \int_{\tau_{0}}^{\tau} E_{\theta}(x_{his},t_{\tau},\tau)),x_{\tau}) \nonumber \\
    \phi^{\ast}=\arg \min_{\phi} \sum_{\tau=\tau_{t-L}}^{\tau_{t}}\mathcal{L}_{r}( D_{\phi}( \int_{\tau_{0}}^{\tau} E_{\theta}(x_{his},t_{\tau},\tau)),x_{\tau})
\end{align}
where $\mathcal{L}_{r}$ and $\mathcal{L}_{p}$ denote the reconstruction and prediction loss, respectively (which will be described in detail in Section~\ref{sec:optimization}). During training, PDETime optimizes both $\theta$ and $\phi$; while during testing, it only optimizes $\phi$ to enhance the extrapolation. To ensure speed and efficiency, we employ a single ridge regression for $D_{\phi}$~\cite{bertinetto2018meta}.



\subsection{Optimization}
\label{sec:optimization}
Different from previous works~\cite{zhou2021informer,zhou2022fedformer, wu2021autoformer,woo2023learning} that directly predict the future time steps with the objective $\mathcal{L}_{p}=\frac{1}{H}  \mathcal{L}(x_{\tau},\hat{x}_{\tau})$. As described as Eq~\ref{eq:pde}, we consider the LMTF as the Initial Value Problem in PDE, and the initial condition $x_{\tau_{0}}$is known (we use the latest time step in the historical series as the initial condition). Thus the objective of PDETime is to estimate the integral term $\int_{\tau_{0}}^{\tau}  \frac{\partial u(x,t_{\tau})}{\partial t_{\tau}} d\tau$. To achieve this, we define the prediction objective as:
\begin{align}
    \mathcal{L}_{p}=\frac{1}{H} \mathcal{L}(D_{\phi}(z_{\tau}),x_{\tau}-x_{\tau_{0}}) 
\end{align}
where $z_{\tau}=\int_{\tau_{0}}^{\tau} E_{\theta}(x_{his},t_{\tau},\tau) d\tau$ as described in Section~\ref{sec:forecaster}. Additionally, $\mathcal{L}_{r}$ has the same form $\mathcal{L}_{r}=\frac{1}{L} (D_{\phi}(z_{\tau}), x_{\tau}-x_{\tau_{0}})$.



In addition to the prediction objective, we also aim to ensure that the first-order difference of the predicted sequence matches that of the target sequence. Therefore, we introduce the optimization objective:
\begin{align}
    \mathcal{L}_{f}=\frac{1}{H} \sum_{\tau=\tau_{t+1}}^{\tau_{t+H}} \mathcal{L}(D_{\phi}(z_{\tau})-D_{\phi}(z_{\tau-1}),x_{\tau}-x_{\tau-1})
\end{align}
Combining $\mathcal{L}_{p}$, $\mathcal{L}_{c}$, and $\mathcal{L}_{f}$, the training objective is:
\begin{align}
\mathcal{L}_{p}=\mathcal{L}_{p}+\mathcal{L}_{c}+\mathcal{L}_{f}.
\end{align}
\section{Experiments}
\subsection{Experimental Settings}
\textbf{Datasets.} We extensively include 7 real-world datasets in our experiments, including four ETT datasets (ETTh1, ETTh2, ETTm1, ETTm2)~\cite{zhou2021informer}. Electricity, Weather and Traffic~\cite{wu2021autoformer}, covering energy, transportation and weather domains (See Appendix~\ref{sec:dataset_details} for more details on the datasets). To ensure a fair evaluation, we follow the standard protocol of dividing each dataset into the training, validation and testing subsets according to the chronological order. The split ratio is 6:2:2 for the ETT dataset and 7:1:2 for the others~\cite{zhou2021informer,wu2021autoformer}. We set the length of the lookback series as 512 for PatchTST, 336 for DLinear, and 96 for other historical-value-based models. The experimental settings of DeepTime remain consistent with the original settings~\cite{woo2023learning}. 
The prediction length varies in $\{ 96, 192, 336, 720 \}$. 

\textbf{Comparison methods.} We carefully choose 10 well-acknowledged models (9 historical-value-based models and 1 time-index-based model) as our benchmarks, including (1) Transformer-based models: Autoformer~\cite{wu2021autoformer}, FEDformer~\cite{zhou2022fedformer}, Stationary~\cite{liu2022non}, Crossformer~\cite{zhang2022crossformer}, PatchTST~\cite{nie2022time}, and iTransformer~\cite{liu2023itransformer}; (2) Linear-based models: DLinear~\cite{zeng2023transformers} ; (3) CNN-based models: SCINet~\cite{liu2022scinet}, TimesNet~\cite{wu2022timesnet}; (4) Time-index-based model: DeepTime~\cite{woo2023learning}. (See Appendix~\ref{baselines_details} for details of these baselines)

\textbf{Implementation Details.} Our method is trained with the Smooth L1~\cite{girshick2015fast} loss using the ADAM~\cite{kingma2014adam} with the initial learning rate selected from $\{10^{-4},10^{-3}, 5\times 10^{-3} \}$. Batch size is set to 32. All experiments are repeated 3 times with fixed seed 2024, implemented in Pytorch~\cite{paszke2019pytorch} and conducted on a single NVIDIA RTX 3090 GPUs. Following DeepTime~\cite{woo2023learning}, we set the look-back length as $L=\mu \ast H$, where $\mu$ is a multiplier which decides the length of the look-back windows. We search through the values $\mu =[1,3,5,7,9]$, and select the best value based on the validation loss.
We set layers of INRs $k=5$ by default, and select the best results from $N=\{ 1,2,3,5\}$. We summarize the temporal features used in this work in Appendix~\ref{sec:time_features}.

\subsection{Main Results and Ablation Study}

\begin{table*}[!]
  \caption{\update{Full results of the long-term forecasting task}. We compare extensive competitive models under different prediction lengths following the setting of PatchTST~\citeyearpar{nie2022time}. The input sequence length is set to 336 and 512 for DLinear and PatchTST, and 96 for other historical-value-based baselines. \emph{Avg} means the average results from all four prediction lengths.}\label{tab:full_baseline_results}
  \vskip -0.0in
  \vspace{3pt}
  \renewcommand{\arraystretch}{0.85} 
  \centering
  \begin{threeparttable}
  \begin{small}
  \renewcommand{\multirowsetup}{\centering}
  \setlength{\tabcolsep}{2.5pt}
  \begin{tabular}{c|c|cc|cc|cc|cc|cc|cc|cc|cc|cc|cc|cc}
    \toprule
    \multicolumn{2}{c}{\multirow{2}{*}{Models}} & 
    \multicolumn{2}{c}{\rotatebox{0}{\scalebox{0.8}{\textbf{PDETime}}}} &
    \multicolumn{2}{c}{\rotatebox{0}{\scalebox{0.8}{{iTransformer}}}} &
    \multicolumn{2}{c}{\rotatebox{0}{\scalebox{0.8}{{PatchTST}}}} &
    \multicolumn{2}{c}{\rotatebox{0}{\scalebox{0.8}{Crossformer}}}  &
    \multicolumn{2}{c}{\rotatebox{0}{\scalebox{0.8}{{DeepTime}}}} &
    \multicolumn{2}{c}{\rotatebox{0}{\scalebox{0.8}{{TimesNet}}}} &
    \multicolumn{2}{c}{\rotatebox{0}{\scalebox{0.8}{{DLinear}}}}&
    \multicolumn{2}{c}{\rotatebox{0}{\scalebox{0.8}{SCINet}}} &
    \multicolumn{2}{c}{\rotatebox{0}{\scalebox{0.8}{FEDformer}}} &
    \multicolumn{2}{c}{\rotatebox{0}{\scalebox{0.8}{Stationary}}} &
    \multicolumn{2}{c}{\rotatebox{0}{\scalebox{0.8}{Autoformer}}} \\
    \multicolumn{2}{c}{} &
    \multicolumn{2}{c}{\scalebox{0.8}{\textbf{(Ours)}}} & 
    \multicolumn{2}{c}{\scalebox{0.8}{\citeyearpar{liu2023itransformer}}} & 
    \multicolumn{2}{c}{\scalebox{0.8}{\citeyearpar{nie2022time}}} & 
    \multicolumn{2}{c}{\scalebox{0.8}{\citeyearpar{zhang2022crossformer}}}  & 
    \multicolumn{2}{c}{\scalebox{0.8}{\citeyearpar{woo2023learning}}} & 
    \multicolumn{2}{c}{\scalebox{0.8}{\citeyearpar{wu2022timesnet}}} & 
    \multicolumn{2}{c}{\scalebox{0.8}{\citeyearpar{zeng2023transformers}}}& 
    \multicolumn{2}{c}{\scalebox{0.8}{\citeyearpar{liu2022scinet}}} &
    \multicolumn{2}{c}{\scalebox{0.8}{\citeyearpar{zhou2022fedformer}}} &
    \multicolumn{2}{c}{\scalebox{0.8}{\citeyearpar{liu2022non}}} &
    \multicolumn{2}{c}{\scalebox{0.8}{\citeyearpar{wu2021autoformer}}} \\
    \cmidrule(lr){3-4} \cmidrule(lr){5-6}\cmidrule(lr){7-8} \cmidrule(lr){9-10}\cmidrule(lr){11-12}\cmidrule(lr){13-14} \cmidrule(lr){15-16} \cmidrule(lr){17-18} \cmidrule(lr){19-20} \cmidrule(lr){21-22} \cmidrule(lr){23-24}
    \multicolumn{2}{c}{Metric}  & \scalebox{0.78}{MSE} & \scalebox{0.78}{MAE}  & \scalebox{0.78}{MSE} & \scalebox{0.78}{MAE}  & \scalebox{0.78}{MSE} & \scalebox{0.78}{MAE}  & \scalebox{0.78}{MSE} & \scalebox{0.78}{MAE}  & \scalebox{0.78}{MSE} & \scalebox{0.78}{MAE}  & \scalebox{0.78}{MSE} & \scalebox{0.78}{MAE} & \scalebox{0.78}{MSE} & \scalebox{0.78}{MAE} & \scalebox{0.78}{MSE} & \scalebox{0.78}{MAE} & \scalebox{0.78}{MSE} & \scalebox{0.78}{MAE} & \scalebox{0.78}{MSE} & \scalebox{0.78}{MAE} & \scalebox{0.78}{MSE} & \scalebox{0.78}{MAE} \\
    \toprule
    
    \multirow{5}{*}{\rotatebox{90}{\scalebox{0.95}{ETTm1}}}
    &  \scalebox{0.78}{96} & \textbf{\scalebox{0.78}{0.292}} & \textbf{\scalebox{0.78}{0.335}} & \scalebox{0.78}{0.334} & \scalebox{0.78}{0.368} & \underline{\scalebox{0.78}{0.293}} & {\scalebox{0.78}{0.346}} & \scalebox{0.78}{0.404} & \scalebox{0.78}{0.426} & \scalebox{0.78}{0.305} & \scalebox{0.78}{0.347} &{\scalebox{0.78}{0.338}} &{\scalebox{0.78}{0.375}} &{\scalebox{0.78}{0.299}} &\underline{\scalebox{0.78}{0.343}} & \scalebox{0.78}{0.418} & \scalebox{0.78}{0.438} &\scalebox{0.78}{0.379} &\scalebox{0.78}{0.419} &\scalebox{0.78}{0.386} &\scalebox{0.78}{0.398} &\scalebox{0.78}{0.505} &\scalebox{0.78}{0.475} \\ 
    & \scalebox{0.78}{192} & \textbf{\scalebox{0.78}{0.329}} & \textbf{\scalebox{0.78}{0.359}} & \scalebox{0.78}{0.377} & \scalebox{0.78}{0.391} & \underline{\scalebox{0.78}{0.333}} & {\scalebox{0.78}{0.370}} & \scalebox{0.78}{0.450} & \scalebox{0.78}{0.451} &\scalebox{0.78}{0.340} & \scalebox{0.78}{0.371} &{\scalebox{0.78}{0.374}} &{\scalebox{0.78}{0.387}}  &{\scalebox{0.78}{0.335}} &\underline{\scalebox{0.78}{0.365}} & \scalebox{0.78}{0.439} & \scalebox{0.78}{0.450}  &\scalebox{0.78}{0.426} &\scalebox{0.78}{0.441} &\scalebox{0.78}{0.459} &\scalebox{0.78}{0.444} &\scalebox{0.78}{0.553} &\scalebox{0.78}{0.496} \\ 
    & \scalebox{0.78}{336} & \textbf{\scalebox{0.78}{0.346}} & \textbf{\scalebox{0.78}{0.374}} & \scalebox{0.78}{0.426} & \scalebox{0.78}{0.420} & {\scalebox{0.78}{0.369}} & {\scalebox{0.78}{0.392}} & \scalebox{0.78}{0.532}  &\scalebox{0.78}{0.515} & \underline{\scalebox{0.78}{0.362}} & \scalebox{0.78}{0.387} &{\scalebox{0.78}{0.410}} &{\scalebox{0.78}{0.411}}  &{\scalebox{0.78}{0.369}} &\underline{\scalebox{0.78}{0.386}} & \scalebox{0.78}{0.490} & \scalebox{0.78}{0.485}  &\scalebox{0.78}{0.445} &\scalebox{0.78}{0.459} &\scalebox{0.78}{0.495} &\scalebox{0.78}{0.464} &\scalebox{0.78}{0.621} &\scalebox{0.78}{0.537} \\ 
    & \scalebox{0.78}{720} & \textbf{\scalebox{0.78}{0.395}} & \textbf{\scalebox{0.78}{0.404}} & \scalebox{0.78}{0.491} & \scalebox{0.78}{0.459} & {\scalebox{0.78}{0.416}} & {\scalebox{0.78}{0.420}} & \scalebox{0.78}{0.666} & \scalebox{0.78}{0.589} & \underline{\scalebox{0.78}{0.399}} & \underline{\scalebox{0.78}{0.414}} &{\scalebox{0.78}{0.478}} &{\scalebox{0.78}{0.450}} &{\scalebox{0.78}{0.425}} &{\scalebox{0.78}{0.421}} & \scalebox{0.78}{0.595} & \scalebox{0.78}{0.550}  &\scalebox{0.78}{0.543} &\scalebox{0.78}{0.490} &\scalebox{0.78}{0.585} &\scalebox{0.78}{0.516} &\scalebox{0.78}{0.671} &\scalebox{0.78}{0.561} \\ 
    \cmidrule(lr){2-24}
    & \scalebox{0.78}{Avg} & \textbf{\scalebox{0.78}{0.340}} & \textbf{\scalebox{0.78}{0.368}} & \scalebox{0.78}{0.407} & \scalebox{0.78}{0.410} & {\scalebox{0.78}{0.352}} & {\scalebox{0.78}{0.382}} & \scalebox{0.78}{0.513} & \scalebox{0.78}{0.496} & \underline{\scalebox{0.78}{0.351}} & \scalebox{0.78}{0.379} &{\scalebox{0.78}{0.400}} &{\scalebox{0.78}{0.406}}  &{\scalebox{0.78}{0.357}} &\underline{\scalebox{0.78}{0.378}} & \scalebox{0.78}{0.485} & \scalebox{0.78}{0.481}  &\scalebox{0.78}{0.448} &\scalebox{0.78}{0.452} &\scalebox{0.78}{0.481} &\scalebox{0.78}{0.456} &\scalebox{0.78}{0.588} &\scalebox{0.78}{0.517} \\ 
    \midrule
    
    \multirow{5}{*}{\update{\rotatebox{90}{\scalebox{0.95}{ETTm2}}}}
    &  \scalebox{0.78}{96} & \textbf{\scalebox{0.78}{0.158}} & \textbf{\scalebox{0.78}{0.244}} & \scalebox{0.78}{0.180} & \scalebox{0.78}{0.264} & \underline{\scalebox{0.78}{0.166}} & \underline{{\scalebox{0.78}{0.256}}} & \scalebox{0.78}{0.287} & \scalebox{0.78}{0.366} & \underline{\scalebox{0.78}{0.166}} & \scalebox{0.78}{0.257} &{\scalebox{0.78}{0.187}} &\scalebox{0.78}{0.267} &\scalebox{0.78}{0.167} &\scalebox{0.78}{0.260} & \scalebox{0.78}{0.286} & \scalebox{0.78}{0.377} &\scalebox{0.78}{0.203} &\scalebox{0.78}{0.287} &{\scalebox{0.78}{0.192}} &\scalebox{0.78}{0.274} &\scalebox{0.78}{0.255} &\scalebox{0.78}{0.339} \\ 
    & \scalebox{0.78}{192} & \textbf{\scalebox{0.78}{0.213}} & \textbf{\scalebox{0.78}{0.283}} & {\scalebox{0.78}{0.250}} & {\scalebox{0.78}{0.309}} & \underline{\scalebox{0.78}{0.223}} & \underline{\scalebox{0.78}{0.296}} & \scalebox{0.78}{0.414} & \scalebox{0.78}{0.492} & \scalebox{0.78}{0.225} & \scalebox{0.78}{0.302} &{\scalebox{0.78}{0.249}} &{\scalebox{0.78}{0.309}} &\scalebox{0.78}{0.224} &\scalebox{0.78}{0.303} & \scalebox{0.78}{0.399} & \scalebox{0.78}{0.445} &\scalebox{0.78}{0.269} &\scalebox{0.78}{0.328} &\scalebox{0.78}{0.280} &\scalebox{0.78}{0.339} &\scalebox{0.78}{0.281} &\scalebox{0.78}{0.340} \\ 
    & \scalebox{0.78}{336} & \textbf{\scalebox{0.78}{0.262}} & \textbf{\scalebox{0.78}{0.318}} & {\scalebox{0.78}{0.311}} & {\scalebox{0.78}{0.348}} & \underline{\scalebox{0.78}{0.274}} & \underline{\scalebox{0.78}{0.329}}  & \scalebox{0.78}{0.597} & \scalebox{0.78}{0.542}  & \scalebox{0.78}{0.277} & \scalebox{0.78}{0.336} &{\scalebox{0.78}{0.321}} &{\scalebox{0.78}{0.351}} &\scalebox{0.78}{0.281} &\scalebox{0.78}{0.342} & \scalebox{0.78}{0.637} & \scalebox{0.78}{0.591} &\scalebox{0.78}{0.325} &\scalebox{0.78}{0.366} &\scalebox{0.78}{0.334} &\scalebox{0.78}{0.361} &\scalebox{0.78}{0.339} &\scalebox{0.78}{0.372} \\ 
    & \scalebox{0.78}{720} & \textbf{\scalebox{0.78}{0.334}} & \textbf{\scalebox{0.78}{0.336}} & {\scalebox{0.78}{0.412}} & {\scalebox{0.78}{0.407}} & \underline{\scalebox{0.78}{0.362}} & \underline{\scalebox{0.78}{0.385}} & \scalebox{0.78}{1.730} & \scalebox{0.78}{1.042} & \scalebox{0.78}{0.383} & \scalebox{0.78}{0.409} &{\scalebox{0.78}{0.408}} &{\scalebox{0.78}{0.403}} &\scalebox{0.78}{0.397} &\scalebox{0.78}{0.421} & \scalebox{0.78}{0.960} & \scalebox{0.78}{0.735} &\scalebox{0.78}{0.421} &\scalebox{0.78}{0.415} &\scalebox{0.78}{0.417} &\scalebox{0.78}{0.413} &\scalebox{0.78}{0.433} &\scalebox{0.78}{0.432} \\ 
    \cmidrule(lr){2-24}
    & \scalebox{0.78}{Avg} & \textbf{\scalebox{0.78}{0.241}} & \textbf{\scalebox{0.78}{0.295}} & {\scalebox{0.78}{0.288}} & {\scalebox{0.78}{0.332}} & \underline{\scalebox{0.78}{0.256}} & \underline{\scalebox{0.78}{0.316}} & \scalebox{0.78}{0.757} & \scalebox{0.78}{0.610} & \scalebox{0.78}{0.262} & \scalebox{0.78}{0.326} &{\scalebox{0.78}{0.291}} &{\scalebox{0.78}{0.333}} &\scalebox{0.78}{0.267} &\scalebox{0.78}{0.331} & \scalebox{0.78}{0.571} & \scalebox{0.78}{0.537} &\scalebox{0.78}{0.305} &\scalebox{0.78}{0.349} &\scalebox{0.78}{0.306} &\scalebox{0.78}{0.347} &\scalebox{0.78}{0.327} &\scalebox{0.78}{0.371} \\ 
    \midrule
    
    \multirow{5}{*}{\rotatebox{90}{\update{\scalebox{0.95}{ETTh1}}}}
    &  \scalebox{0.78}{96} & \textbf{\scalebox{0.78}{0.356}} & \textbf{\scalebox{0.78}{0.381}} & \scalebox{0.78}{0.386} & {\scalebox{0.78}{0.405}} & \scalebox{0.78}{0.379} & \scalebox{0.78}{0.401} & \scalebox{0.78}{0.423} & \scalebox{0.78}{0.448} & \underline{\scalebox{0.78}{0.371}}& \underline{\scalebox{0.78}{0.396}}  &{\scalebox{0.78}{0.384}} &{\scalebox{0.78}{0.402}} & \scalebox{0.78}{0.375} &{\scalebox{0.78}{0.399}} & \scalebox{0.78}{0.654} & \scalebox{0.78}{0.599} &{\scalebox{0.78}{0.376}} &\scalebox{0.78}{0.419} &\scalebox{0.78}{0.513} &\scalebox{0.78}{0.491} &\scalebox{0.78}{0.449} &\scalebox{0.78}{0.459}  \\ 
    & \scalebox{0.78}{192} & \textbf{\scalebox{0.78}{0.397}} & \textbf{\scalebox{0.78}{0.406}} & {\scalebox{0.78}{0.441}} & {\scalebox{0.78}{0.436}} & \scalebox{0.78}{0.413} & \scalebox{0.78}{0.429} & \scalebox{0.78}{0.471} & \scalebox{0.78}{0.474}  & \underline{\scalebox{0.78}{0.403}} & \underline{\scalebox{0.78}{0.420}} &{\scalebox{0.78}{0.436}} &{\scalebox{0.78}{0.429}}  &{\scalebox{0.78}{0.405}} &{\scalebox{0.78}{0.416}} & \scalebox{0.78}{0.719} & \scalebox{0.78}{0.631} &{\scalebox{0.78}{0.420}} &\scalebox{0.78}{0.448} &\scalebox{0.78}{0.534} &\scalebox{0.78}{0.504} &\scalebox{0.78}{0.500} &\scalebox{0.78}{0.482} \\ 
    & \scalebox{0.78}{336} & \textbf{\scalebox{0.78}{0.420}} & \textbf{\scalebox{0.78}{0.419}} & {\scalebox{0.78}{0.487}} & {\scalebox{0.78}{0.458}} & \scalebox{0.78}{0.435} & \underline{\scalebox{0.78}{0.436}} & \scalebox{0.78}{0.570} & \scalebox{0.78}{0.546} & \underline{\scalebox{0.78}{0.433}} & \underline{\scalebox{0.78}{0.436}} &\scalebox{0.78}{0.491} &\scalebox{0.78}{0.469} &{\scalebox{0.78}{0.439}} & {\scalebox{0.78}{0.443}} & \scalebox{0.78}{0.778} & \scalebox{0.78}{0.659} & {\scalebox{0.78}{0.459}} &{\scalebox{0.78}{0.465}} &\scalebox{0.78}{0.588} &\scalebox{0.78}{0.535} &\scalebox{0.78}{0.521} &\scalebox{0.78}{0.496} \\ 
    & \scalebox{0.78}{720} & \textbf{\scalebox{0.78}{0.425}} & \textbf{\scalebox{0.78}{0.446}} & {\scalebox{0.78}{0.503}} & 
    {\scalebox{0.78}{0.491}} & \underline{\scalebox{0.78}{0.446}} & \underline{\scalebox{0.78}{0.464}} & \scalebox{0.78}{0.653} & \scalebox{0.78}{0.621} & \scalebox{0.78}{0.474} & \scalebox{0.78}{0.492} &\scalebox{0.78}{0.521} &{\scalebox{0.78}{0.500}} &\scalebox{0.78}{0.472} &\scalebox{0.78}{0.490} & \scalebox{0.78}{0.836} & \scalebox{0.78}{0.699} &{\scalebox{0.78}{0.506}} &{\scalebox{0.78}{0.507}} &\scalebox{0.78}{0.643} &\scalebox{0.78}{0.616} &{\scalebox{0.78}{0.514}} &\scalebox{0.78}{0.512}  \\ 
    \cmidrule(lr){2-24}
    & \scalebox{0.78}{Avg} & \textbf{\scalebox{0.78}{0.399}} & \textbf{\scalebox{0.78}{0.413}} & {\scalebox{0.78}{0.454}} & {\scalebox{0.78}{0.447}} & \underline{\scalebox{0.78}{0.418}} & \underline{\scalebox{0.78}{0.432}} & \scalebox{0.78}{0.529} & \scalebox{0.78}{0.522} & \scalebox{0.78}{0.420} & \scalebox{0.78}{0.436} &\scalebox{0.78}{0.458} &{\scalebox{0.78}{0.450}} &{\scalebox{0.78}{0.423}} &{\scalebox{0.78}{0.437}} & \scalebox{0.78}{0.747} & \scalebox{0.78}{0.647} &{\scalebox{0.78}{0.440}} &\scalebox{0.78}{0.460} &\scalebox{0.78}{0.570} &\scalebox{0.78}{0.537} &\scalebox{0.78}{0.496} &\scalebox{0.78}{0.487}  \\ 
    \midrule

    \multirow{5}{*}{\rotatebox{90}{\scalebox{0.95}{ETTh2}}}
    &  \scalebox{0.78}{96} & \textbf{\scalebox{0.78}{0.268}} & \textbf{\scalebox{0.78}{0.330}} & {\scalebox{0.78}{0.297}} & {\scalebox{0.78}{0.349}} & \underline{\scalebox{0.78}{0.274}} & \underline{\scalebox{0.78}{0.335}} & \scalebox{0.78}{0.745} & \scalebox{0.78}{0.584} &\scalebox{0.78}{0.287} & \scalebox{0.78}{0.352}  & {\scalebox{0.78}{0.340}} & {\scalebox{0.78}{0.374}} &{\scalebox{0.78}{0.289}} &{\scalebox{0.78}{0.353}} & \scalebox{0.78}{0.707} & \scalebox{0.78}{0.621}  &\scalebox{0.78}{0.358} &\scalebox{0.78}{0.397} &\scalebox{0.78}{0.476} &\scalebox{0.78}{0.458} &\scalebox{0.78}{0.346} &\scalebox{0.78}{0.388} \\ 
    & \scalebox{0.78}{192} & \textbf{\scalebox{0.78}{0.331}} & \textbf{\scalebox{0.78}{0.370}} & {\scalebox{0.78}{0.380}} & {\scalebox{0.78}{0.400}} &\underline{\scalebox{0.78}{0.342}} & \underline{\scalebox{0.78}{0.382}} & \scalebox{0.78}{0.877} & \scalebox{0.78}{0.656} & \scalebox{0.78}{0.383} & \scalebox{0.78}{0.412} & {\scalebox{0.78}{0.402}} & {\scalebox{0.78}{0.414}} &\scalebox{0.78}{0.383} &\scalebox{0.78}{0.418} & \scalebox{0.78}{0.860} & \scalebox{0.78}{0.689} &{\scalebox{0.78}{0.429}} &{\scalebox{0.78}{0.439}} &\scalebox{0.78}{0.512} &\scalebox{0.78}{0.493} &\scalebox{0.78}{0.456} &\scalebox{0.78}{0.452} \\ 
    & \scalebox{0.78}{336} & \textbf{\scalebox{0.78}{0.358}} & \textbf{\scalebox{0.78}{0.395}} & {\scalebox{0.78}{0.428}} & {\scalebox{0.78}{0.432}} & \underline{\scalebox{0.78}{0.365}} & \underline{\scalebox{0.78}{0.404}}& \scalebox{0.78}{1.043} & \scalebox{0.78}{0.731} & \scalebox{0.78}{0.523} & \scalebox{0.78}{0.501}  & {\scalebox{0.78}{0.452}} & {\scalebox{0.78}{0.452}} &\scalebox{0.78}{0.448} &\scalebox{0.78}{0.465} & \scalebox{0.78}{1.000} &\scalebox{0.78}{0.744} &\scalebox{0.78}{0.496} &\scalebox{0.78}{0.487} &\scalebox{0.78}{0.552} &\scalebox{0.78}{0.551} &{\scalebox{0.78}{0.482}} &\scalebox{0.78}{0.486}\\ 
    & \scalebox{0.78}{720} & \textbf{\scalebox{0.78}{0.380}} & \textbf{\scalebox{0.78}{0.421}} & {\scalebox{0.78}{0.427}} & {\scalebox{0.78}{0.445}} & \underline{\scalebox{0.78}{0.393}} & \underline{\scalebox{0.78}{0.430}} & \scalebox{0.78}{1.104} & \scalebox{0.78}{0.763} & \scalebox{0.78}{0.765} & \scalebox{0.78}{0.624} & {\scalebox{0.78}{0.462}} & {\scalebox{0.78}{0.468}} &\scalebox{0.78}{0.605} &\scalebox{0.78}{0.551} & \scalebox{0.78}{1.249} & \scalebox{0.78}{0.838} &{\scalebox{0.78}{0.463}} &{\scalebox{0.78}{0.474}} &\scalebox{0.78}{0.562} &\scalebox{0.78}{0.560} &\scalebox{0.78}{0.515} &\scalebox{0.78}{0.511} \\ 
    \cmidrule(lr){2-24}
    & \scalebox{0.78}{Avg} & \textbf{\scalebox{0.78}{0.334}} & \textbf{\scalebox{0.78}{0.379}} & {\scalebox{0.78}{0.383}} & {\scalebox{0.78}{0.407}} & \underline{\scalebox{0.78}{0.343}} & \underline{\scalebox{0.78}{0.387}} & \scalebox{0.78}{0.942} & \scalebox{0.78}{0.684} & \scalebox{0.78}{0.489} & \scalebox{0.78}{0.472}  &{\scalebox{0.78}{0.414}} &{\scalebox{0.78}{0.427}} &\scalebox{0.78}{0.431} &\scalebox{0.78}{0.446} & \scalebox{0.78}{0.954} & \scalebox{0.78}{0.723} &\scalebox{0.78}{{0.437}} &\scalebox{0.78}{{0.449}} &\scalebox{0.78}{0.526} &\scalebox{0.78}{0.516} &\scalebox{0.78}{0.450} &\scalebox{0.78}{0.459} \\ 
    \midrule
    
    \multirow{5}{*}{\rotatebox{90}{\scalebox{0.95}{ECL}}} 
    &  \scalebox{0.78}{96} & \textbf{\scalebox{0.78}{0.129}} & \textbf{\scalebox{0.78}{0.222}} & \scalebox{0.78}{0.148} & \scalebox{0.78}{0.240} & \textbf{\scalebox{0.78}{0.129}} & \textbf{\scalebox{0.78}{0.222}} & \scalebox{0.78}{0.219} & \scalebox{0.78}{0.314} & \scalebox{0.78}{0.137} & \scalebox{0.78}{0.238} &{\scalebox{0.78}{0.168}} &{\scalebox{0.78}{0.272}} &\scalebox{0.78}{0.140} &\scalebox{0.78}{0.237} & \scalebox{0.78}{0.247} & \scalebox{0.78}{0.345} &\scalebox{0.78}{0.193} &\scalebox{0.78}{0.308} &{\scalebox{0.78}{0.169}} &{\scalebox{0.78}{0.273}} &\scalebox{0.78}{0.201} &\scalebox{0.78}{0.317}  \\ 
    & \scalebox{0.78}{192} & \textbf{\scalebox{0.78}{0.143}} & \textbf{\scalebox{0.78}{0.235}} & \scalebox{0.78}{0.162} & {\scalebox{0.78}{0.253}} & \underline{\scalebox{0.78}{0.147}} & \underline{\scalebox{0.78}{0.240}} & \scalebox{0.78}{0.231} & \scalebox{0.78}{0.322} & \scalebox{0.78}{0.152} & \scalebox{0.78}{0.252} &{\scalebox{0.78}{0.184}} &\scalebox{0.78}{0.289} &\scalebox{0.78}{0.153} &{\scalebox{0.78}{0.249}} & \scalebox{0.78}{0.257} & \scalebox{0.78}{0.355} &\scalebox{0.78}{0.201} &\scalebox{0.78}{0.315} &{\scalebox{0.78}{0.182}} &\scalebox{0.78}{0.286} &\scalebox{0.78}{0.222} &\scalebox{0.78}{0.334} \\ 
    & \scalebox{0.78}{336} & \textbf{\scalebox{0.78}{0.152}} & \textbf{\scalebox{0.78}{0.248}} & \scalebox{0.78}{0.178} & {\scalebox{0.78}{0.269}} & \underline{\scalebox{0.78}{0.163}} & \underline{\scalebox{0.78}{0.259}} & \scalebox{0.78}{0.246} & \scalebox{0.78}{0.337} & \scalebox{0.78}{0.166} & \scalebox{0.78}{0.268} &{\scalebox{0.78}{0.198}} &{\scalebox{0.78}{0.300}} &\scalebox{0.78}{0.169} &{\scalebox{0.78}{0.267}} & \scalebox{0.78}{0.269} & \scalebox{0.78}{0.369} &\scalebox{0.78}{0.214} &\scalebox{0.78}{0.329} &{\scalebox{0.78}{0.200}} &\scalebox{0.78}{0.304} &\scalebox{0.78}{0.231} &\scalebox{0.78}{0.338}  \\ 
    & \scalebox{0.78}{720} & \textbf{\scalebox{0.78}{0.176}} & \textbf{\scalebox{0.78}{0.272}} & \scalebox{0.78}{0.225} & \scalebox{0.78}{0.317} & \underline{\scalebox{0.78}{0.197}} & \underline{\scalebox{0.78}{0.290}} & \scalebox{0.78}{0.280} & \scalebox{0.78}{0.363} & \scalebox{0.78}{0.201} & \scalebox{0.78}{0.302} &{\scalebox{0.78}{0.220}} &{\scalebox{0.78}{0.320}} &\scalebox{0.78}{0.203} &\scalebox{0.78}{0.301} & \scalebox{0.78}{0.299} & \scalebox{0.78}{0.390} &\scalebox{0.78}{0.246} &\scalebox{0.78}{0.355} &{\scalebox{0.78}{0.222}} &{\scalebox{0.78}{0.321}} &\scalebox{0.78}{0.254} &\scalebox{0.78}{0.361} \\ 
    \cmidrule(lr){2-24}
    & \scalebox{0.78}{Avg} & \textbf{\scalebox{0.78}{0.150}} & \textbf{\scalebox{0.78}{0.244}} & \scalebox{0.78}{0.178} & \scalebox{0.78}{0.270} & \underline{\scalebox{0.78}{0.159}} & \underline{\scalebox{0.78}{0.252}} & \scalebox{0.78}{0.244} & \scalebox{0.78}{0.334} & \scalebox{0.78}{0.164} & \scalebox{0.78}{0.265} &{\scalebox{0.78}{0.192}} &{\scalebox{0.78}{0.295}} &\scalebox{0.78}{0.166} &\scalebox{0.78}{0.263} & \scalebox{0.78}{0.268} & \scalebox{0.78}{0.365} &\scalebox{0.78}{0.214} &\scalebox{0.78}{0.327} &{\scalebox{0.78}{0.193}} &{\scalebox{0.78}{0.296}} &\scalebox{0.78}{0.227} &\scalebox{0.78}{0.338} \\ 
    \midrule

    \multirow{5}{*}{\rotatebox{90}{\scalebox{0.95}{Traffic}}} 
    & \scalebox{0.78}{96} & \textbf{\scalebox{0.78}{0.330}} & \textbf{\scalebox{0.78}{0.232}} & \scalebox{0.78}{0.395} & \scalebox{0.78}{0.268} & \underline{\scalebox{0.78}{0.360}} & \underline{\scalebox{0.78}{0.249}} & {\scalebox{0.78}{0.522}} & {\scalebox{0.78}{0.290}} & \scalebox{0.78}{0.390} & \scalebox{0.78}{0.275} &{\scalebox{0.78}{0.593}} &{\scalebox{0.78}{0.321}} &\scalebox{0.78}{0.410} &\scalebox{0.78}{0.282} & \scalebox{0.78}{0.788} & \scalebox{0.78}{0.499} &{\scalebox{0.78}{0.587}} &\scalebox{0.78}{0.366} &\scalebox{0.78}{0.612} &{\scalebox{0.78}{0.338}} &\scalebox{0.78}{0.613} &\scalebox{0.78}{0.388} \\ 
    & \scalebox{0.78}{192} & \textbf{\scalebox{0.78}{0.332}} & \textbf{\scalebox{0.78}{0.232}} & \scalebox{0.78}{0.417} & \scalebox{0.78}{0.276} & \underline{\scalebox{0.78}{0.379}} & \underline{\scalebox{0.78}{0.256}} & {\scalebox{0.78}{0.530}} & {\scalebox{0.78}{0.293}} & \scalebox{0.78}{0.402} & \scalebox{0.78}{0.278} &\scalebox{0.78}{0.617} &{\scalebox{0.78}{0.336}} &{\scalebox{0.78}{0.423}} &\scalebox{0.78}{0.287} & \scalebox{0.78}{0.789} & \scalebox{0.78}{0.505} &\scalebox{0.78}{0.604} &\scalebox{0.78}{0.373} &\scalebox{0.78}{0.613} &{\scalebox{0.78}{0.340}} &\scalebox{0.78}{0.616} &\scalebox{0.78}{0.382}  \\ 
    & \scalebox{0.78}{336} & \textbf{\scalebox{0.78}{0.342}} & \textbf{\scalebox{0.78}{0.236}} & \scalebox{0.78}{0.433} & \scalebox{0.78}{0.283} & \underline{\scalebox{0.78}{0.392}} & \underline{\scalebox{0.78}{0.264}} & \scalebox{0.78}{0.558} & {\scalebox{0.78}{0.305}}  & \scalebox{0.78}{0.415} & \scalebox{0.78}{0.288} &\scalebox{0.78}{0.629} &{\scalebox{0.78}{0.336}}  &{\scalebox{0.78}{0.436}} &\scalebox{0.78}{0.296} & \scalebox{0.78}{0.797} & \scalebox{0.78}{0.508}&\scalebox{0.78}{0.621} &\scalebox{0.78}{0.383} &\scalebox{0.78}{0.618} &{\scalebox{0.78}{0.328}} &\scalebox{0.78}{0.622} &\scalebox{0.78}{0.337} \\ 
    & \scalebox{0.78}{720} & \textbf{\scalebox{0.78}{0.365}} & \textbf{\scalebox{0.78}{0.244}} & \scalebox{0.78}{0.467} & \scalebox{0.78}{0.302} & \underline{\scalebox{0.78}{0.432}} & \underline{\scalebox{0.78}{0.286}} & \scalebox{0.78}{0.589} & {\scalebox{0.78}{0.328}}  & \scalebox{0.78}{0.449} & \scalebox{0.78}{0.307} &\scalebox{0.78}{0.640} &{\scalebox{0.78}{0.350}} &\scalebox{0.78}{0.466} &\scalebox{0.78}{0.315} & \scalebox{0.78}{0.841} & \scalebox{0.78}{0.523} &{\scalebox{0.78}{0.626}} &\scalebox{0.78}{0.382} &\scalebox{0.78}{0.653} &{\scalebox{0.78}{0.355}} &\scalebox{0.78}{0.660} &\scalebox{0.78}{0.408} \\ 
    \cmidrule(lr){2-24}
    & \scalebox{0.78}{Avg} & \textbf{\scalebox{0.78}{0.342}} & \textbf{\scalebox{0.78}{0.236}} & \scalebox{0.78}{0.428} & \scalebox{0.78}{0.282} & \underline{\scalebox{0.78}{0.390}} & \underline{\scalebox{0.78}{0.263}} & {\scalebox{0.78}{0.550}} & {\scalebox{0.78}{0.304}} & \scalebox{0.78}{0.414} & \scalebox{0.78}{0.287} &{\scalebox{0.78}{0.620}} &{\scalebox{0.78}{0.336}} &\scalebox{0.78}{0.433} &\scalebox{0.78}{0.295} & \scalebox{0.78}{0.804} & \scalebox{0.78}{0.509} &{\scalebox{0.78}{0.610}} &\scalebox{0.78}{0.376} &\scalebox{0.78}{0.624} &{\scalebox{0.78}{0.340}} &\scalebox{0.78}{0.628} &\scalebox{0.78}{0.379} \\ 
    \midrule
    
    \multirow{5}{*}{\rotatebox{90}{\scalebox{0.95}{Weather}}} 
    &  \scalebox{0.78}{96} & \underline{\scalebox{0.78}{0.157}} & \underline{\scalebox{0.78}{0.203}} & \scalebox{0.78}{0.174} & \scalebox{0.78}{0.214} & \textbf{\scalebox{0.78}{0.149}} & \textbf{\scalebox{0.78}{0.198}} & {\scalebox{0.78}{0.158}} & \scalebox{0.78}{0.230}  & \scalebox{0.78}{0.166} & \scalebox{0.78}{0.221} &{\scalebox{0.78}{0.172}} &{\scalebox{0.78}{0.220}} & \scalebox{0.78}{0.176} &\scalebox{0.78}{0.237} & \scalebox{0.78}{0.221} & \scalebox{0.78}{0.306} & \scalebox{0.78}{0.217} &\scalebox{0.78}{0.296} & {\scalebox{0.78}{0.173}} &{\scalebox{0.78}{0.223}} & \scalebox{0.78}{0.266} &\scalebox{0.78}{0.336} \\ 
    & \scalebox{0.78}{192} & \underline{\scalebox{0.78}{0.200}} & \underline{\scalebox{0.78}{0.246}} & \scalebox{0.78}{0.221} & \scalebox{0.78}{0.254} & \textbf{\scalebox{0.78}{0.194}} & \textbf{\scalebox{0.78}{0.241}} & {\scalebox{0.78}{0.206}} & \scalebox{0.78}{0.277} & \scalebox{0.78}{0.207} & \scalebox{0.78}{0.261} &{\scalebox{0.78}{0.219}} &{\scalebox{0.78}{0.261}}  & \scalebox{0.78}{0.220} &\scalebox{0.78}{0.282} & \scalebox{0.78}{0.261} & \scalebox{0.78}{0.340} & \scalebox{0.78}{0.276} &\scalebox{0.78}{0.336} & \scalebox{0.78}{0.245} &\scalebox{0.78}{0.285} & \scalebox{0.78}{0.307} &\scalebox{0.78}{0.367} \\ 
    & \scalebox{0.78}{336} & \textbf{\scalebox{0.78}{0.241}} & \textbf{\scalebox{0.78}{0.281}} & \scalebox{0.78}{0.278} & \scalebox{0.78}{0.296} & \underline{\scalebox{0.78}{0.245}} & \underline{\scalebox{0.78}{0.282}} & {\scalebox{0.78}{0.272}} & \scalebox{0.78}{0.335} & \scalebox{0.78}{0.251} & \scalebox{0.78}{0.298} &{\scalebox{0.78}{0.280}} &{\scalebox{0.78}{0.306}} & \scalebox{0.78}{0.265} &\scalebox{0.78}{0.319} & \scalebox{0.78}{0.309} & \scalebox{0.78}{0.378} & \scalebox{0.78}{0.339} &\scalebox{0.78}{0.380} & \scalebox{0.78}{0.321} &\scalebox{0.78}{0.338} & \scalebox{0.78}{0.359} &\scalebox{0.78}{0.395}\\ 
    & \scalebox{0.78}{720} & \textbf{\scalebox{0.78}{0.291}} & \textbf{\scalebox{0.78}{0.324}} & \scalebox{0.78}{0.358} & \scalebox{0.78}{0.349} & \underline{\scalebox{0.78}{0.314}} & \underline{\scalebox{0.78}{0.334}} & \scalebox{0.78}{0.398} & \scalebox{0.78}{0.418} & {\scalebox{0.78}{0.301}} & \scalebox{0.78}{0.338} &\scalebox{0.78}{0.365} &{\scalebox{0.78}{0.359}} & {\scalebox{0.78}{0.323}} &{\scalebox{0.78}{0.362}} & \scalebox{0.78}{0.377} & \scalebox{0.78}{0.427} & \scalebox{0.78}{0.403} &\scalebox{0.78}{0.428} & \scalebox{0.78}{0.414} &\scalebox{0.78}{0.410} & \scalebox{0.78}{0.419} &\scalebox{0.78}{0.428} \\ 
    \cmidrule(lr){2-24}
    & \scalebox{0.78}{Avg} & \textbf{\scalebox{0.78}{0.222}} & \textbf{\scalebox{0.78}{0.263}} & \scalebox{0.78}{0.258} & \scalebox{0.78}{0.279} & \underline{\scalebox{0.78}{0.225}} & \underline{\scalebox{0.78}{0.263}} & \scalebox{0.78}{0.259} & \scalebox{0.78}{0.315} & \scalebox{0.78}{0.231} & \scalebox{0.78}{0.286} &{\scalebox{0.78}{0.259}} &{\scalebox{0.78}{0.287}} &\scalebox{0.78}{0.246} &\scalebox{0.78}{0.300} & \scalebox{0.78}{0.292} & \scalebox{0.78}{0.363} &\scalebox{0.78}{0.309} &\scalebox{0.78}{0.360} &\scalebox{0.78}{0.288} &\scalebox{0.78}{0.314} &\scalebox{0.78}{0.338} &\scalebox{0.78}{0.382} \\ 
    \midrule

      \multicolumn{2}{c|}{\scalebox{0.78}{{$1^{\text{st}}$ Count}}} & \scalebox{0.78}{{33}} & \scalebox{0.78}{{33}} & \scalebox{0.78}{0} & \scalebox{0.78}{{0}} & \scalebox{0.78}{{2}} & \scalebox{0.78}{2} & \scalebox{0.78}{0} & \scalebox{0.78}{0} & \scalebox{0.78}{0} & \scalebox{0.78}{0} & \scalebox{0.78}{0} & \scalebox{0.78}{0} & \scalebox{0.78}{0} & \scalebox{0.78}{0} & \scalebox{0.78}{0} & \scalebox{0.78}{0} & \scalebox{0.78}{0} & \scalebox{0.78}{0} & \scalebox{0.78}{0} & \scalebox{0.78}{0} & \scalebox{0.78}{0} & \scalebox{0.78}{0} \\ 
    \bottomrule
  \end{tabular}
    \end{small}
  \end{threeparttable}
\label{tab:main_result}
\end{table*}
Comprehensive forecasting results are listed in Table~\ref{tab:main_result} with the best in \textbf{Bold} and the second \underline{underlined}. The lower MSE/MAE indicates the more accurate prediction result. Overall, PDETime achieves the best performance on most settings across seven real-world datasets compared with historical-value-based and time-index-based models. Additionally, experimental results also show that the performance of the proposed PDETime changes quite steadily as the prediction length $H$ increases. For instance, the MSE of PDETime increases from 0.330 to 0.365 on the Traffic dataset, while the MSE of PatchTST increases from 0.360 to 0.432, which is the SOTA historical-value-based model. This phenomenon was observed in other datasets and settings as well, indicating that PDETime retains better long-term robustness, which is meaningful for real-world practical applications.


\begin{table*}[!]
\caption{Ablation study on variants of PDETime. -Temporal refers that removing the temporal domain feature $t_{\tau}$; -Spatial refers that removing the historical observations $x_{his}$; - Initial  refers that removing the initial condition $x_{\tau_{0}}$. The best results are highlighted in \textbf{bold}.}
    \centering
     \vskip -0.0in
    \vspace{3pt}
    \centering
    \setlength{\tabcolsep}{3pt}
    \begin{threeparttable}
    \begin{small}
        
    \begin{tabular}{c|c|cc| cc|cc| cc|cc| ccc}
    \toprule
    \multirow{2}{*}{\scalebox{0.8}{Dataset}} & \scalebox{0.8}{Models} & \multicolumn{2}{c|}{\scalebox{0.8}{PDETime}} & \multicolumn{2}{c|}{\scalebox{0.8}{-Temporal}} & \multicolumn{2}{c|}{\scalebox{0.8}{-Spatial}} & \multicolumn{2}{c|}{\scalebox{0.8}{-Initial }} & \multicolumn{2}{c|}{\scalebox{0.9}{-Temporal -Spatial}}  & \multicolumn{2}{c}{\scalebox{0.9}{- All}} \\
     & \scalebox{0.8}{Metric} & \scalebox{0.8}{MSE} & \scalebox{0.8}{MAE} & \scalebox{0.8}{MSE} & \scalebox{0.8}{MAE} & \scalebox{0.8}{MSE} & \scalebox{0.8}{MAE} & \scalebox{0.8}{MSE} & \scalebox{0.8}{MAE} & \scalebox{0.8}{MSE} & \scalebox{0.8}{MAE} & \scalebox{0.8}{MSE} & \scalebox{0.8}{MAE}  \\
     \midrule
     \multirow{4}{*}{\scalebox{0.9}{Traffic}} &\scalebox{0.8}{96} & \scalebox{0.8}{0.330} & \scalebox{0.8}{\textbf{0.232}} & \scalebox{0.8}{0.336} & \scalebox{0.8}{0.236} & \scalebox{0.8}{\textbf{0.329}} & \scalebox{0.8}{\textbf{0.232}} & \scalebox{0.8}{0.334} & \scalebox{0.8}{0.235} & \scalebox{0.8}{0.394} & \scalebox{0.8}{0.268} & \scalebox{0.8}{0.401} & \scalebox{0.8}{0.269}  \\
     &\scalebox{0.8}{192} & \scalebox{0.8}{\textbf{0.332}} & \scalebox{0.8}{\textbf{0.232}} & \scalebox{0.8}{0.368} & \scalebox{0.8}{0.247} & \scalebox{0.8}{0.336} & \scalebox{0.8}{0.234} & \scalebox{0.8}{0.334} & \scalebox{0.8}{\textbf{0.232}} & \scalebox{0.8}{0.407} & \scalebox{0.8}{0.269} & \scalebox{0.8}{0.413} & \scalebox{0.8}{0.270}  \\
     &\scalebox{0.8}{336} & \scalebox{0.8}{\textbf{0.342}} & \scalebox{0.8}{\textbf{0.236}} & \scalebox{0.8}{0.378} & \scalebox{0.8}{0.251} & \scalebox{0.8}{0.344} & \scalebox{0.8}{\textbf{0.236}} & \scalebox{0.8}{0.343} & \scalebox{0.8}{\textbf{0.236}} & \scalebox{0.8}{0.419} & \scalebox{0.8}{0.273} & \scalebox{0.8}{0.426} & \scalebox{0.8}{0.272}  \\
     &\scalebox{0.8}{720} & \scalebox{0.8}{\textbf{0.365}} & \scalebox{0.8}{\textbf{0.244}} & \scalebox{0.8}{0.406} & \scalebox{0.8}{0.265} & \scalebox{0.8}{0.371} & \scalebox{0.8}{0.250} & \scalebox{0.8}{0.368} & \scalebox{0.8}{0.250} & \scalebox{0.8}{0.453} & \scalebox{0.8}{0.291} & \scalebox{0.8}{0.671} & \scalebox{0.8}{0.406}  \\
     \midrule
     \multirow{4}{*}{\scalebox{0.9}{Weather}} &\scalebox{0.8}{96} & \scalebox{0.8}{\textbf{0.157}} & \scalebox{0.8}{\textbf{0.203}} & \scalebox{0.8}{0.158} & \scalebox{0.8}{0.205} & \scalebox{0.8}{0.159} & \scalebox{0.8}{0.205} & \scalebox{0.8}{0.169} & \scalebox{0.8}{0.213} & \scalebox{0.8}{0.159} & \scalebox{0.8}{0.205} & \scalebox{0.8}{0.166} & \scalebox{0.8}{0.212}  \\
     &\scalebox{0.8}{192} & \scalebox{0.8}{0.200} & \scalebox{0.8}{0.246} & \scalebox{0.8}{0.206} & \scalebox{0.8}{0.253} & \scalebox{0.8}{\textbf{0.198}} & \scalebox{0.8}{\textbf{0.243}} & \scalebox{0.8}{0.208} & \scalebox{0.8}{0.248} & \scalebox{0.8}{\textbf{0.198}} & \scalebox{0.8}{\textbf{0.243}} & \scalebox{0.8}{0.208} & \scalebox{0.8}{0.250}  \\
     &\scalebox{0.8}{336} & \scalebox{0.8}{0.241} & \scalebox{0.8}{0.281} & \scalebox{0.8}{\textbf{0.240}} & \scalebox{0.8}{0.278} & \scalebox{0.8}{0.246} & \scalebox{0.8}{0.282} & \scalebox{0.8}{0.245} & \scalebox{0.8}{0.287} & \scalebox{0.8}{\textbf{0.240}} & \scalebox{0.8}{\textbf{0.277}} & \scalebox{0.8}{0.244} & \scalebox{0.8}{0.283}  \\
     &\scalebox{0.8}{720} & \scalebox{0.8}{0.291} & \scalebox{0.8}{0.324} & \scalebox{0.8}{0.292} & \scalebox{0.8}{0.323} & \scalebox{0.8}{\textbf{0.290}} & \scalebox{0.8}{\textbf{0.322}} & \scalebox{0.8}{0.300} & \scalebox{0.8}{0.337} & \scalebox{0.8}{0.294} & \scalebox{0.8}{0.327} & \scalebox{0.8}{0.299} & \scalebox{0.8}{0.337}  \\
     \bottomrule
    \end{tabular}
    \end{small}
    \end{threeparttable}
    \label{tab:ablation}
\end{table*}
We perform ablation studies on the Traffic and Weather datasets to validate the effect of \textbf{temporal feature} $t_{\tau}$, \textbf{spatial feature} $x_{his}$ and \textbf{initial condition} $x_{\tau_{0}}$. The results are presented in Table~\ref{tab:ablation}. 
\textbf{1)} The initial condition $x_{t_{0}}$ is useful on most settings. As mentioned in Section~\ref{sec:approach}, we treat LMTF as Initial Value Problem, thus the effectiveness of $x_{\tau_{0}}$ validates the correctness of PDETime. 
\textbf{2)} The impact of spatial features $x_{his}$ on PDETime is limited. This may be due to the fact that the true spatial domains $x$ are unknown, and we utilize the historical observations $x_{his}$ to simulate $x$. However, this approach may not be effective in capturing the true spatial features.
\textbf{3)} The influence of temporal feature $t_{\tau}$ on PDETime various significantly across different datasets. 
Experimental results have shown that $t_{\tau}$ is highly beneficial in the Traffic dataset, but its effect on Weather dataset is limited. 
For example, the period of Traffic dataset may be one day or one week, making it easier for PDETime to learn temporal features. On the other hand, the period of Weather dataset may be one year or longer, but the dataset only contains one year of data. As a result, PDETime cannot capture the complete temporal features in this case.
\par
We conduct an additional ablation study on Traffic to evaluate the ability of different INRs to extract features of $x_{his}$, $t_{\tau}$, and $\tau$. In this study, we compared the performance of using the GELU or Tanh activation function instead of sine in SIREN and making $\tau^{(0)}=[\text{GELU}(2\pi \textbf{B}_{1}\tau), \text{GELU}(2\pi \textbf{B}_{1}\tau),...]$ or $\tau^{(0)}=[\text{Tanh}(2\pi \textbf{B}_{1}\tau), \text{Tanh}(2\pi \textbf{B}_{1}\tau),...]$.
Table~\ref{tab:act} presents the experimental results, we observe that the sine function (periodic functions) can extract features better than other non-decreasing activation functions. This is because the smooth, non-periodic activation functions fail to accurately model high-frequency information~\cite{sitzmann2020implicit}. Time series data is often periodic, and the periodic nature of the sine function makes it more effective in extracting time series features.

\subsection{Effects of Hyper-parameters}
\begin{figure*}[!]
\vskip 0.2in
\begin{center}
\centerline{\includegraphics[width=0.92\textwidth-0.2in]{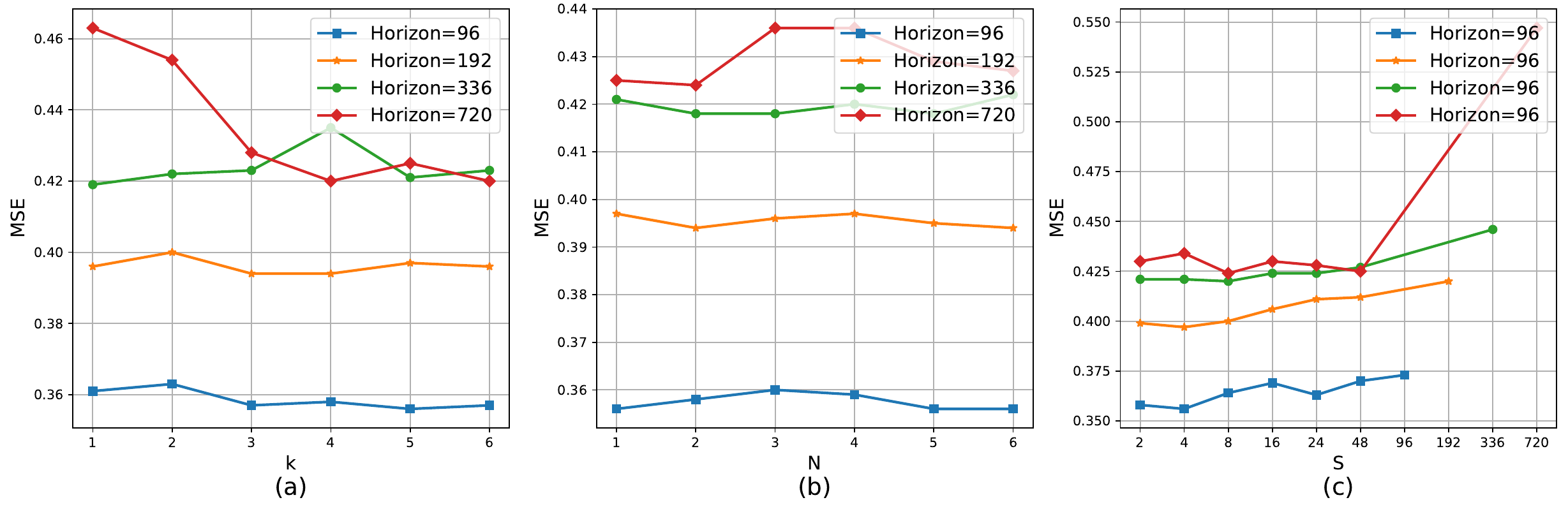}}\vspace{-5mm}
\caption{Evaluation on hyper-parameter impact. (a) MSE against hyper-parameter layers of INRs $k$ in Forecaster on ETTh1. (b) MSE against hyper-parameter layers of aggregation module $N$ in Forecaster on ETTh1. (c) MSE against hyper-parameter patch length $S$ in Estimator on ETTh1. }
\label{fig:params}
\end{center}\vspace{-11mm}
\end{figure*}
We evaluate the effect of four hyper-parameters: look-back window $L$, number of INRs layers $k$, number of aggregation layers $N$, and patch length $S$ on the ETTh1 and ETTh2 datasets. First, we perform a sensitivity on the look-back window $L=\mu \ast H$, where $H$ is the experimental setting. 
The results are presented in Table~\ref{tab:mu}. 
We observe that the test error decreases as $\mu$ increases, plateauing and even increasing slightly as $\mu$ grows extremely large when the horizon window is small. 
However, under a large horizon window, the test error increases as $\mu$ increases.
Next, we evaluate the hyper-parameters $N$ and $k$ on PDETime, as shown in Figure~\ref{fig:params} (a) and (b) respectively. We find that the performance of PDETime remains stable when $k \geq 3$. Additionally, the number of aggregation layers $N$ has a limited impact on PDETime.
Furthermore, we investigate the effect of patch length $S$ on PDETime, as illustrated in Figure~\ref{fig:params} (c). We varied the patch length from 2 to 48 and evaluate MSE with different horizon windows. As the patch length $S$ increased, the prediction accuracy of PDETime initially improved, reached a peak, and then started to decline. However, the accuracy remains relatively stable throughout. We also extended the patch length to $S=H$. In this case, PDETime performed poorly, indicating that the accumulation of errors has a significant impact on the performance of PDETime.
Overall, these analyses provide insights into the effects of different hyper-parameters on the performance of PDETime and can guide the selection of appropriate settings for achieving optimal results.

Due to space limitations, we put extra experiments on robustness and visualization experiments in  Appendix~\ref{sec:sup_results} and Appendix~\ref{sec:vis}, respectively.



\begin{table}[!t]
\caption{Analysis on the look-back window length, based on the multiplier on horizon length, $L=\mu \ast H$. The best results are highlighted in \textbf{bold}.}
    \vskip -0.0in
    \vspace{3pt}
    \centering
    \setlength{\tabcolsep}{3pt}
    \begin{threeparttable}  
    \begin{small}
        
    \begin{tabular}{cccccccccc}
    \toprule
    \multirow{2}{*}{\scalebox{0.9}{Dataset}} & \scalebox{0.9}{Horizon} & \multicolumn{2}{c}{\scalebox{0.9}{96}} & \multicolumn{2}{c}{\scalebox{0.9}{192}} & \multicolumn{2}{c}{\scalebox{0.9}{336}} & \multicolumn{2}{c}{\scalebox{0.9}{720}} \\
      & \scalebox{0.9}{ $\mu$ } & \scalebox{0.8}{MSE} & \scalebox{0.8}{MAE} & \scalebox{0.8}{MSE} & \scalebox{0.8}{MAE} & \scalebox{0.8}{MSE} & \scalebox{0.8}{MAE} & \scalebox{0.8}{MSE} & \scalebox{0.8}{MAE} \\
     \midrule 
     \multirow{5}{*}{ETTh1} & \scalebox{0.9}{ 1 } & \scalebox{0.8}{0.378} & \scalebox{0.8}{0.386} & \scalebox{0.8}{0.415} & \scalebox{0.8}{0.411} & \scalebox{0.8}{\textbf{0.421}} & \scalebox{0.8}{\textbf{0.420}} & \scalebox{0.8}{\textbf{0.425}} & \scalebox{0.8}{\textbf{0.446}} \\
      & \scalebox{0.9}{ 3 } & \scalebox{0.8}{0.359} & \scalebox{0.8}{0.382} & \scalebox{0.8}{\textbf{0.394}} & \scalebox{0.8}{\textbf{0.404}} & \scalebox{0.8}{0.427} & \scalebox{0.8}{0.421} & \scalebox{0.8}{0.443} & \scalebox{0.8}{0.460} \\
      & \scalebox{0.9}{ 5 } & \scalebox{0.8}{0.360} & \scalebox{0.8}{0.385} & \scalebox{0.8}{0.396} & \scalebox{0.8}{0.405} & \scalebox{0.8}{\textbf{0.421}} & \scalebox{0.8}{\textbf{0.420}} & \scalebox{0.8}{0.495} & \scalebox{0.8}{0.501} \\
      & \scalebox{0.9}{ 7 } & \scalebox{0.8}{\textbf{0.354}} & \scalebox{0.8}{\textbf{0.381}} & \scalebox{0.8}{0.398} & \scalebox{0.8}{0.405} & \scalebox{0.8}{0.427} & \scalebox{0.8}{0.429} & \scalebox{0.8}{0.545} & \scalebox{0.8}{0.532} \\
      & \scalebox{0.9}{ 9 } & \scalebox{0.8}{0.356} & \scalebox{0.8}{\textbf{0.381}} & \scalebox{0.8}{0.397} & \scalebox{0.8}{0.406} & \scalebox{0.8}{0.446} & \scalebox{0.8}{0.440} & \scalebox{0.8}{1.220} & \scalebox{0.8}{0.882} \\
      \midrule
      \multirow{5}{*}{ETTh2} & \scalebox{0.9}{ 1 } & \scalebox{0.8}{0.288} & \scalebox{0.8}{0.335} & \scalebox{0.8}{0.357} & \scalebox{0.8}{0.381} & \scalebox{0.8}{0.380} & \scalebox{0.8}{0.404} & \scalebox{0.8}{\textbf{0.380}} & \scalebox{0.8}{\textbf{0.421}} \\
      & \scalebox{0.9}{ 3 } & \scalebox{0.8}{0.276} & \scalebox{0.8}{0.331} & \scalebox{0.8}{0.339} & \scalebox{0.8}{0.374} & \scalebox{0.8}{\textbf{0.358}} & \scalebox{0.8}{\textbf{0.395}} & \scalebox{0.8}{0.422} & \scalebox{0.8}{0.456} \\
      & \scalebox{0.9}{ 5 } & \scalebox{0.8}{0.275} & \scalebox{0.8}{0.333} & \scalebox{0.8}{\textbf{0.331}} & \scalebox{0.8}{\textbf{0.370}} & \scalebox{0.8}{0.360} & \scalebox{0.8}{0.408} & \scalebox{0.8}{0.622} & \scalebox{0.8}{0.576} \\
      & \scalebox{0.9}{ 7 } & \scalebox{0.8}{\textbf{0.268}} & \scalebox{0.8}{\textbf{0.330}} & \scalebox{0.8}{\textbf{0.331}} & \scalebox{0.8}{0.378} & \scalebox{0.8}{0.384} & \scalebox{0.8}{0.427} & \scalebox{0.8}{0.624} & \scalebox{0.8}{0.595} \\
      & \scalebox{0.9}{ 9 } & \scalebox{0.8}{0.272} & \scalebox{0.8}{0.331} & \scalebox{0.8}{\textbf{0.331}} & \scalebox{0.8}{0.378} & \scalebox{0.8}{0.412} & \scalebox{0.8}{0.451} & \scalebox{0.8}{0.797} & \scalebox{0.8}{0.689} \\
      \bottomrule
      
    \end{tabular}
    \end{small}
    \end{threeparttable}\vspace{-3mm}
    \label{tab:mu}
\end{table}

\begin{table}[t]
\caption{Analysis on INRs. PDETime refers to our proposed approach. GELU and Tanh refer to replacing SIREN and CFF with GELU or Tanh activation, respectively. The best results are highlighted in \textbf{bold}.}
    \vskip -0.0in
    \vspace{3pt}
    \centering
    \setlength{\tabcolsep}{3.5pt}
    \begin{threeparttable}  
    \begin{small}
        
    \begin{tabular}{cccccccc}
    \toprule
    \multirow{2}{*}{\scalebox{1.0}{Dataset}} & \scalebox{1.0}{Method} & \multicolumn{2}{c}{\scalebox{1.0}{PDETime}} & \multicolumn{2}{c}{\scalebox{1.0}{GELU}} & \multicolumn{2}{c}{\scalebox{1.0}{Tanh}} \\
      & \scalebox{1.0}{ Metric } & \scalebox{1.0}{MSE} & \scalebox{1.0}{MAE} & \scalebox{1.0}{MSE} & \scalebox{1.0}{MAE} & \scalebox{1.0}{MSE} & \scalebox{1.0}{MAE}  \\
     \midrule 
     \multirow{4}{*}{Traffic} & \scalebox{1.0}{ 96 } & \scalebox{1.0}{\textbf{0.330}} & \scalebox{1.0}{\textbf{0.232}} & \scalebox{1.0}{0.332} & \scalebox{1.0}{0.237} & \scalebox{1.0}{{0.338}} & \scalebox{1.0}{{0.233}} \\
      & \scalebox{1.0}{ 192 } & \scalebox{1.0}{\textbf{0.332}} & \scalebox{1.0}{\textbf{0.232}} & \scalebox{1.0}{{0.338}} & \scalebox{1.0}{{0.241}} & \scalebox{1.0}{0.339} & \scalebox{1.0}{0.235} \\
      & \scalebox{1.0}{ 336 } & \scalebox{1.0}{\textbf{0.342}} & \scalebox{1.0}{\textbf{0.236}} & \scalebox{1.0}{0.348} & \scalebox{1.0}{0.244} & \scalebox{1.0}{{0.348}} & \scalebox{1.0}{{0.238}}  \\
      & \scalebox{1.0}{ 720 } & \scalebox{1.0}{\textbf{0.365}} & \scalebox{1.0}{\textbf{0.244}} & \scalebox{1.0}{0.376} & \scalebox{1.0}{0.252} & \scalebox{1.0}{0.366} & \scalebox{1.0}{\textbf{0.244}} \\
      \bottomrule
      
    \end{tabular}
    \end{small}
    \end{threeparttable}\vspace{-3mm}
    \label{tab:act}
\end{table}

\section{Conclusion and Future Work}
In this paper, we propose a novel LMTS framework PDETime, based on neural PDE solvers, which consists of Encoder, Solver, and Decoder. Specifically, the Encoder simulates the partial differentiation of the original function over time at each time step in latent space in parallel. 
The solver is responsible for computing the integral term with improved stability. 
Finally, the Decoder maps the integral term from latent space into the spatial function and predicts the target series given the initial condition. 
Additionally, we incorporate meta-optimization techniques to enhance the ability of PDETime to extrapolate future series.
Extensive experimental results show that PDETime achieves state-of-the-art performance across forecasting benchmarks on various real-world datasets. We also perform ablation studies to identify the key components contributing to the success of PDETime. 


\textbf{Future Work} Firstly, while our proposed neural PDE solver, PDETime, has shown promising results for long-term multivariate time series forecasting, there are other types of neural PDE solvers that could potentially be applied to this task. Exploring these alternative neural PDE solvers and comparing their performance on LMTF could be an interesting future direction. 
Furthermore, our ablation studies have revealed that the effectiveness of the historical observations (spatial domain) on PDETime is limited, and the impact of the temporal domain varies significantly across different datasets. Designing effective network structures that better utilize the spatial and temporal domains could be an important area for future research. 
Finally, our approach of rethinking long-term multivariate time series forecasting from the perspective of partial differential equations has led to state-of-the-art performance. Exploring other perspectives and frameworks to tackle this task could be a promising direction for future research.

 \section*{Broader Impacts}
 Our research is dedicated to innovating time series forecasting techniques to push the boundaries of time series analysis further. While our primary objective is to enhance predictive accuracy and efficiency, we are also mindful of the broader ethical considerations that accompany technological progress in this area. While immediate societal impacts may not be apparent, we acknowledge the importance of ongoing vigilance regarding the ethical use of these advancements. It is essential to continuously assess and address potential implications to ensure responsible development and application in various sectors.
\label{submission}

\balance
\bibliography{example_paper}
\bibliographystyle{icml2024}

\newpage
\appendix
\onecolumn
\section{Appendix}
In this section, we present the experimental details of PDETime. The organization of this section is as follows:
\begin{itemize}
    \item Appendix~\ref{sec:exp_detail} provides details on the datasets and baselines.
    \item Appendix~\ref{sec:time_features} provides details of time features used in this work.
    \item Appendix~\ref{sec:code_Encoder} provides pseudocode of the Encoder of PDETime.
    \item Appendix~\ref{sec:code_Solver} provides pseudocode of the Solver of PDETime.
    \item Appendix~\ref{sec:sup_results} presents the results of the robustness experiments.
    \item Appendix~\ref{sec:vis} visualizes the prediction results of PDETime on the Traffic dataset.
\end{itemize}

\subsection{Experimental Details}
\label{sec:exp_detail}
\subsubsection{Datasets}
\label{sec:dataset_details}
we use the most popular multivariate datasets in LMTF, including ETT, Electricity, Traffic and Weather:
\begin{itemize}
    \item The ETT~\citep{zhou2021informer} (Electricity Transformer Temperature) dataset contains two years of data from two separate countries in China with intervals of 1-hour level (ETTh) and 15-minute level (ETTm) collected from electricity transformers. Each time step contains six power load features and oil temperature.
    \item The Electricity \footnote[1]{ \href{https://archive.ics.uci.edu/ml/datasets/ElectricityLoadDiagrams20112014}{https://archive.ics.uci.edu/ml/datasets/ElectricityLoadDiagrams20112014.}} dataset describes 321 clients' hourly electricity consumption from 2012 to 2014.
    \item The Traffic \footnote[2]{\href{http://pems.dot.ca.gov}{http://pems.dot.ca.gov.}} dataset contains the road occupancy rates from various sensors on San Francisco Bay area freeways, which is provided by California Department of Transportation.
    \item the Weather \footnote[3]{\href{https://www.bgc-jena.mpg.de/wetter/}{https://www.bgc-jena.mpg.de/wetter/.}} dataset contains 21 meteorological indicators collected at around 1,600 landmarks in the United States.
\end{itemize}
Table~\ref{tab:dataset_char} presents key characteristics of the seven datasets. The dimensions of each dataset range from 7 to 862, with frequencies ranging from 10 minutes to 7 days. The length of the datasets varies from 966 to 69,680 data points. We split all datasets into training, validation, and test sets in chronological order, using a ratio of 6:2:2 for the ETT dataset and 7:1:2 for the remaining datasets.

\begin{table*}[htbp]
    \centering
    \caption{Statics of Dataset Characteristics.}
    \begin{threeparttable}
    \begin{tabular}{cccccccc}
    \toprule
    Datasets & ETTh1 & ETTh2 & ETTm1 & ETTm2 & Electricity & Traffic & Weather \\
    \midrule
    Dimension & 7 & 7 & 7 & 7 & 321 & 862 & 21 \\
    Frequency & 1 hour & 1 hour & 15 min & 15 min & 1 hour & 1 hour & 10 min \\
    Length &  17420 & 17420 & 69680 & 69680 & 26304 & 52696 & 17544 \\
    \bottomrule
    
    \end{tabular}
    \end{threeparttable}
    \label{tab:dataset_char}
\end{table*}

\subsubsection{Baselines}
\label{baselines_details}
We choose SOTA and the most representative LMTF models as our baselines, including historical-value-based and time-index-based models, as follows:
\begin{itemize}
    \item PatchTST~\cite{nie2022time}: the current historical-value-based SOTA models. It utilizes channel-independent and patch techniques and achieves the highest performance by utilizing the native Transformer.
    \item DLinear~\cite{zeng2023transformers}: a highly insightful work that employs simple linear models and trend decomposition techniques, outperforming all Transformer-based models at the time.
    \item Crossformer~\cite{zhang2022crossformer}:  similar to PatchTST, it utilizes the patch technique commonly used in the CV domain. However, unlike PatchTST’s independent channel design, it leverages cross-dimension dependency to enhance LMTF performance.
    \item FEDformer~\cite{zhou2022fedformer}: it employs trend decomposition and Fourier transformation techniques to improve the performance of Transformer-based models in LMTF. It was the best-performing Transformer-based model before Dlinear.
    \item Autoformer~\cite{wu2021autoformer}: it combines trend decomposition techniques with an auto-correlation mechanism, inspiring subsequent work such as FEDformer.
    \item Stationary~\cite{liu2022non}: it proposes a De-stationary Attention to alleviate the over-stationarization problem.
    \item iTransformer~\cite{liu2023itransformer}: it different from pervious works that embed multivariate points of each time step as a (temporal) token, it embeds the whole time series of each variate independently into a (variate) token, which is the extreme case of Patching.
    \item TimesNet~\cite{wu2022timesnet}: it transforms the 1D time series into a set of 2D tensors based on multiple periods and uses a parameter-efficient inception block to analyze time series.
    \item SCINet~\cite{liu2022scinet}: it proposes a recursive downsample-convolve-interact architecture to aggregate multiple resolution features with complex temporal dynamics. 
    \item DeepTime~\cite{woo2023learning}: it is the first time-index-based model in long-term multivariate time-series forecasting.
\end{itemize}

\subsection{Temporal Features}
\label{sec:time_features}

Depending on the sampling frequency, the temporal feature $t_{\tau}$ of each dataset is also different. We will introduce the temporal feature of each data set in detail:
\begin{itemize}
    \item ETTm and Weather: day-of-year, month-of-year, day-of-week, hour-of-day, minute-of-hour.
    \item ETTh, Traffic, and Electricity: day-of-year, month-of-year, day-of-week, hour-of-day.
\end{itemize}
we also normalize these features into [0,1] range.


\subsection{Pseudocode}\label{sec:code_Encoder}
We provide the pseudo-code of Encoder and Solver in Algorithms~\ref{alg:1}.

\begin{algorithm}[h]  
	\renewcommand{\algorithmicrequire}{\textbf{Input:}}
	\renewcommand{\algorithmicensure}{\textbf{Output:}}
	\caption{Pseudocode of the aggregation layer of Encoder}  
	\label{alg:1}
	\begin{algorithmic}[1] 
    \Require Time-index feature $\tau$, temporal feature $t_{\tau}$ and spatial feature $x_{his}$.
    \State $\tau$, $t_{\tau}$, $x_{his}$ = Linear($\tau$), Linear($t_{\tau}$), Linear($x_{his}$) \Comment{$\tau \in \mathbb{R}^{B \times L \times d}$, $x_{his} \in \mathbb{R}^{B \times C \times d}$, $t_{\tau} \in \mathbb{R}^{B \times L \times t}$}
    \State $\tau$, $t_{\tau}$, $x_{his}$ = Norm(GeLU($\tau$)), Norm($\sin(t_{\tau})$), Norm(GeLU($x_{his}$))
    \State $\tau$ = Norm($\tau$ + Attn($\tau$, $x_{his}$))
    \State $t_{\tau}$=torch.cat(($\tau$,$t_{\tau}$),dim=-1) \Comment{$t_{\tau} \in \mathbb{R}^{B \times L \times (d+t)}$}
    \State $\tau$=Norm($\tau$ + Linear($t_{\tau}$)) \Comment{$\tau \in \mathbb{R}^{B \times L \times d}$}
    \State \Return{$\tau$}
		
	\end{algorithmic}
\end{algorithm}

\subsection{Pseudocode}\label{sec:code_Solver}

We provide the pseudo-code of the Solver in Algorithms~\ref{alg:2}.

\begin{algorithm}[h]  
	\renewcommand{\algorithmicrequire}{\textbf{Input:}}
	\renewcommand{\algorithmicensure}{\textbf{Output:}}
	\caption{Pseudocode of the numerical Solver}  
	\label{alg:2}
	\begin{algorithmic}[1] 
    \Require time-index feature $\tau \in \mathbb{R}^{B\times L \times d}$, batch size $B$, input length $L$ and dimension $d$, and patch length $S$.
    \State dudt = MLP ($\tau$)  \Comment{dudt $\in \mathbb{R}^{B \times L \times d}$}
    
    \State  u = \text{MLP} ($\tau$)  \Comment{u $\in \mathbb{R}^{B \times L \times d}$}
    
    \State u=u.reshape(B, -1, S, d) \Comment{u $\in \mathbb{R}^{B\times \frac{L}{S}} \times S \times d$}
    \State dudt = -dudt
    \State dudt = torch.flip(dudt, [2])
    \State dudt = dudt\_[:,:,:-1,:] \Comment{dudt $\in \mathbb{R}^{B \times \frac{L}{S} \times (S-1) \times d}$}
    \State dudt = torch.cat((u[:,:,-1:,:], dudt), dim=-2)
    \Comment{dudt $\in \mathbb{R}^{B\times \frac{L}{S}} \times S \times d$}
    \State intdudt = torch.cumsum(dudt, dim=-2)
    \State intdudt = torch.flip(dudt, [2])
    \State intdudt = intdudt.reshape(B, L , d) \Comment{intdudt $\in \mathbb{R}^{B \times L \times d}$}
    \State intdudt = MLP(intdudt) \Comment{intdudt $\in \mathbb{R}^{B \times L \times d}$}
    \State \Return{intdudt}

	\end{algorithmic}
\end{algorithm}

\subsection{Experimental Results of Robustness}
\label{sec:sup_results}

The experimental results of the robustness of our algorithm based on Solver and Initial condition are summarized in Table~\ref{robustness_table}. We also test the effectiveness of continuity loss $\mathcal{L}_{c}$ in Table~\ref{tab:cont}. The experimental results in Table~\ref{robustness_table} demonstrate that PDETime can achieve strong performance on the ETT dataset even when using only the Solver or initial value conditions, without explicitly incorporating spatial and temporal information. This finding suggests that PDETime has the potential to perform well even in scenarios with limited data availability. Additionally, we conducted an analysis on the ETTh1 and ETTh2 datasets to investigate the impact of the loss term $\mathcal{L}_{c}$. Our findings demonstrate that incorporating $\mathcal{L}_{c}$ into PDETime can enhance its robustness.

\begin{table*}[h]
\caption{Analysis on Solver and Initial condition. INRs refers to only using INRs to represent $\tau$; + Initial refers to aggregating initial condition $x_{\tau_{0}}$; +Solver refers to using numerical solvers to compute integral terms in latent space. The best results are highlighted in \textbf{bold}.}
    \vskip -0.0in
    \vspace{3pt}
    \centering
    \setlength{\tabcolsep}{3.5pt}
    \begin{threeparttable}  
    \begin{small}
        
    \begin{tabular}{cccccccc}
    \toprule
    \multirow{2}{*}{\scalebox{1.0}{Dataset}} & \scalebox{1.0}{Method} & \multicolumn{2}{c}{\scalebox{1.0}{INRs}} & \multicolumn{2}{c}{\scalebox{1.0}{INRs+Initial}} & \multicolumn{2}{c}{\scalebox{1.0}{INRs+Solver}} \\
      & \scalebox{1.0}{ Metric } & \scalebox{1.0}{MSE} & \scalebox{1.0}{MAE} & \scalebox{1.0}{MSE} & \scalebox{1.0}{MAE} & \scalebox{1.0}{MSE} & \scalebox{1.0}{MAE}  \\
     \midrule 
     \multirow{4}{*}{ETTh1} & \scalebox{1.0}{ 96 } & \scalebox{1.0}{{0.371}} & \scalebox{1.0}{{0.396}} & \scalebox{1.0}{0.364} & \scalebox{1.0}{0.384} & \scalebox{1.0}{{0.358}} & \scalebox{1.0}{{0.381}} \\
      & \scalebox{1.0}{ 192 } & \scalebox{1.0}{{0.403}} & \scalebox{1.0}{{0.420}} & \scalebox{1.0}{{0.402}} & \scalebox{1.0}{{0.409}} & \scalebox{1.0}{0.398} & \scalebox{1.0}{0.407} \\
      & \scalebox{1.0}{ 336 } & \scalebox{1.0}{{0.433}} & \scalebox{1.0}{{0.436}} & \scalebox{1.0}{0.428} & \scalebox{1.0}{0.420} & \scalebox{1.0}{{0.348}} & \scalebox{1.0}{{0.238}}  \\
      & \scalebox{1.0}{ 720 } & \scalebox{1.0}{{0.474}} & \scalebox{1.0}{{0.492}} & \scalebox{1.0}{0.439} & \scalebox{1.0}{0.452} & \scalebox{1.0}{0.455} & \scalebox{1.0}{{0.476}} \\
      \midrule 
     \multirow{4}{*}{ETTh2} & \scalebox{1.0}{ 96 } & \scalebox{1.0}{{0.287}} & \scalebox{1.0}{{0.352}} & \scalebox{1.0}{0.270} & \scalebox{1.0}{0.330} & \scalebox{1.0}{{0.285}} & \scalebox{1.0}{{0.342}} \\
      & \scalebox{1.0}{ 192 } & \scalebox{1.0}{{0.383}} & \scalebox{1.0}{{0.412}} & \scalebox{1.0}{{0.331}} & \scalebox{1.0}{{0.372}} & \scalebox{1.0}{0.345} & \scalebox{1.0}{0.379} \\
      & \scalebox{1.0}{ 336 } & \scalebox{1.0}{{0.523}} & \scalebox{1.0}{{0.501}} & \scalebox{1.0}{0.373} & \scalebox{1.0}{0.405} & \scalebox{1.0}{{0.357}} & \scalebox{1.0}{{0.399}}  \\
      & \scalebox{1.0}{ 720 } & \scalebox{1.0}{{0.765}} & \scalebox{1.0}{{0.624}} & \scalebox{1.0}{0.392} & \scalebox{1.0}{0.429} & \scalebox{1.0}{0.412} & \scalebox{1.0}{{0.444}} \\
      \bottomrule
      
    \end{tabular}
    \end{small}
    \end{threeparttable}
    \label{robustness_table}
\end{table*}

\begin{table*}[!h]
\caption{Analysis on the effectiveness of loss term $\mathcal{L}_{c}$. }
    \vskip -0.0in
    \vspace{3pt}
    \centering
    \setlength{\tabcolsep}{3.5pt}
    \begin{threeparttable}  
    \begin{small}
        
    \begin{tabular}{cccccc}
    \toprule
    \multirow{2}{*}{\scalebox{1.0}{Dataset}} & \scalebox{1.0}{Method} & \multicolumn{2}{c}{\scalebox{1.0}{PDETime}} & \multicolumn{2}{c}{\scalebox{1.0}{PDETime-$\mathcal{L}_{c}$}}  \\
      & \scalebox{1.0}{ Metric } & \scalebox{1.0}{MSE} & \scalebox{1.0}{MAE} & \scalebox{1.0}{MSE} & \scalebox{1.0}{MAE} \\
     \midrule 
     \multirow{4}{*}{ETTh1} & 96 & 0.356 & 0.381 & 0.357 & 0.381 \\
     & 192 & 0.397 & 0.406 & 0.393 & 0.405 \\
     & 336 & 0.420 & 0.419 & 0.422 & 0.420 \\
     & 720 & 0.425 & 0.419 & 0.446 & 0.458 \\
     \midrule
     \multirow{4}{*}{ETTh2} & 96 & 0.268 & 0.330 & 0.271 & 0.330 \\
     & 192 & 0.331 & 0.370 & 0.341 & 0.373 \\
     & 336 & 0.358 & 0.395 & 0.363 & 0.397 \\
     & 720 & 0.380 & 0.421 & 0.396 & 0.434 \\

      \bottomrule
      
    \end{tabular}
    \end{small}
    \end{threeparttable}
    \label{tab:cont}
\end{table*}

\subsection{Visualization} \label{sec:vis}
\begin{figure*}[h!]
\vskip 0.2in
\begin{center}
\centerline{\includegraphics[width=0.92\textwidth-0.5in]{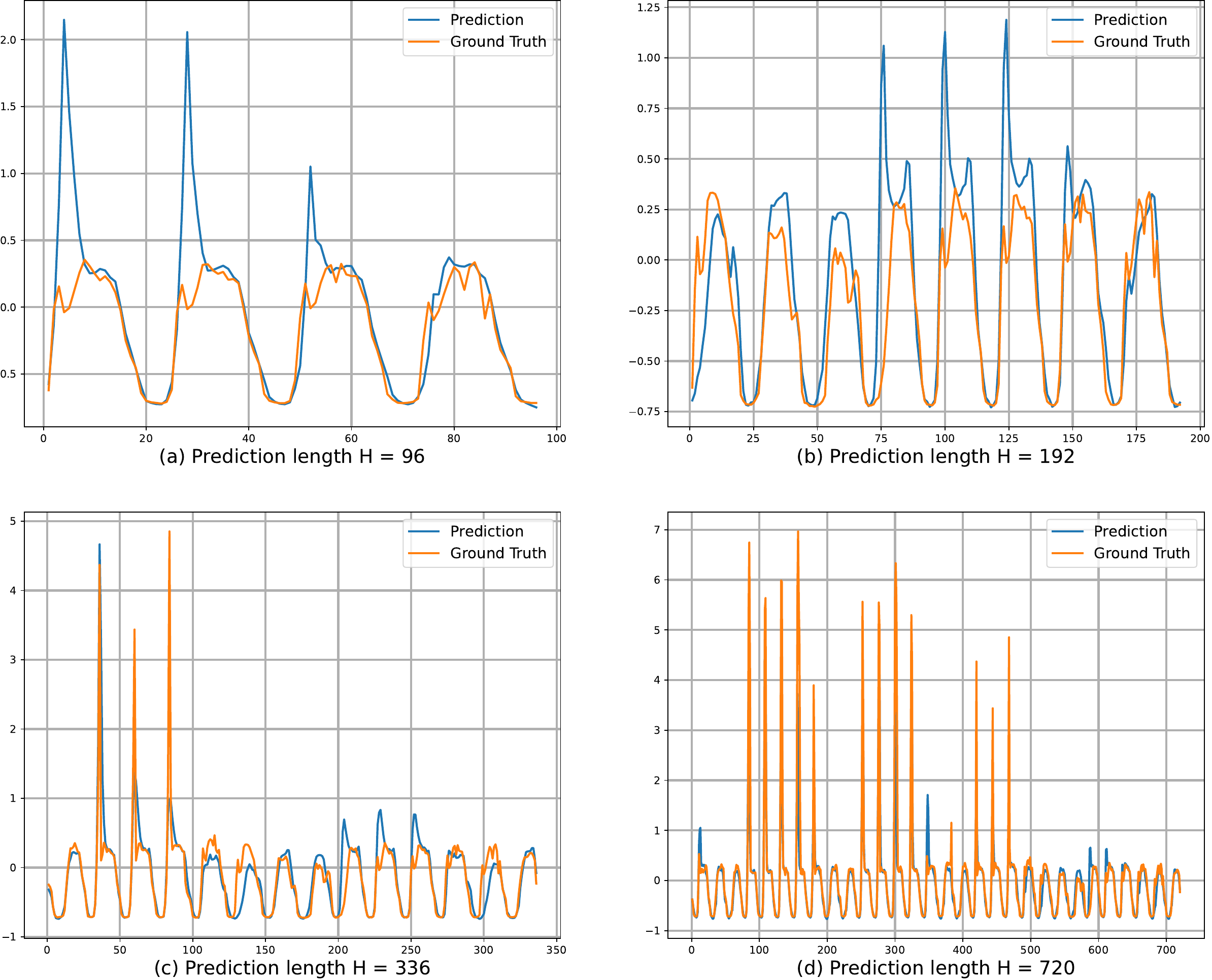}}\vspace{-5mm}
\caption{Visualization of PDETime's prediction results on Traffic with different prediction length. }
\label{fig:vis}
\end{center}\vspace{-11mm}
\end{figure*}

We visualize the prediction results of PDETime on the Traffic dataset. As illustrated in Figure~\ref{fig:vis}, for prediction lengths $H=96, 192, 336, 720$, the prediction curve closely aligns with the ground-truth curves, indicating the outstanding predictive performance of PDETime. Meanwhile, PDETime demonstrates effectiveness in capturing periods of time features. 

\end{document}